\documentclass{article}

\usepackage{microtype}
\usepackage{graphicx}
\usepackage{subfigure}
\usepackage{booktabs} %

\usepackage{hyperref}

\usepackage[accepted]{icml2024}

\usepackage{amsmath}
\usepackage{amssymb}
\usepackage{mathtools}
\usepackage{amsthm}

\usepackage[capitalize,noabbrev]{cleveref}

\theoremstyle{plain}
\newtheorem{theorem}{Theorem}[section]

\newtheorem{lemma}[theorem]{Lemma}

\theoremstyle{definition}

\theoremstyle{remark}
\newtheorem{remark}[theorem]{Remark}

\usepackage[textsize=tiny]{todonotes}

\usepackage{amsmath,amsfonts,bm}

\def\vzero{{\bm{0}}}

\def\vc{{\bm{c}}}
\def\vd{{\bm{d}}}

\def\vf{{\bm{f}}}
\def\vg{{\bm{g}}}
\def\vh{{\bm{h}}}

\def\vm{{\bm{m}}}
\def\vn{{\bm{n}}}

\def\vu{{\bm{u}}}

\def\vx{{\bm{x}}}
\def\vy{{\bm{y}}}

\def\mC{{\bm{C}}}

\def\mF{{\bm{F}}}
\def\mG{{\bm{G}}}
\def\mH{{\bm{H}}}

\def\mS{{\bm{S}}}

\def\mW{{\bm{W}}}

\DeclareMathAlphabet{\mathsfit}{\encodingdefault}{\sfdefault}{m}{sl}
\SetMathAlphabet{\mathsfit}{bold}{\encodingdefault}{\sfdefault}{bx}{n}

\def\gE{{\mathcal{E}}}
\def\gF{{\mathcal{F}}}
\def\gG{{\mathcal{G}}}
\def\gH{{\mathcal{H}}}

\def\gN{{\mathcal{N}}}

\def\gU{{\mathcal{U}}}
\def\gV{{\mathcal{V}}}

\newcommand{\E}{\mathbb{E}}

\newcommand{\R}{\mathbb{R}}

\usepackage{makecell}

\newcommand{\supp}[1]{^{(#1)}}
\newcommand{\sups}[1]{^{[#1]}}
\newcommand{\G}{\mathcal{G}}
\newcommand{\V}{\mathcal{V}}
\newcommand{\VV}{\vert\mathcal{V}\vert}
\renewcommand{\E}{\mathcal{E}}

\newcommand{\Z}{\mathbb{Z}}

\newcommand{\pdiff}[2]{\frac{\partial #1}{\partial #2}}

\newcommand{\En}{{\mathrm{E}(n)}}

\icmltitlerunning{Graph Neural PDE Solvers with Conservation and Similarity-Equivariance}

\begin{document}

\twocolumn[
\icmltitle{Graph Neural PDE Solvers with Conservation and Similarity-Equivariance}

\icmlsetsymbol{equal}{*}

\begin{icmlauthorlist}
\icmlauthor{Masanobu Horie}{ricos}
\icmlauthor{Naoto Mitsume}{tsukuba}
\end{icmlauthorlist}

\icmlaffiliation{ricos}{RICOS Co. Ltd., Tokyo, Japan}
\icmlaffiliation{tsukuba}{Graduate School of Science and Technology, University of Tsukuba, Ibaraki, Japan}

\icmlcorrespondingauthor{Masanobu Horie}{horie@ricos.co.jp}

\icmlkeywords{Graph Neural Network, Physics, Conservation, Equivariance, Partial Differential Equation}

\vskip 0.3in
]

\printAffiliationsAndNotice{}  %

\begin{abstract}
Utilizing machine learning to address partial differential equations (PDEs) presents significant challenges due to the diversity of spatial domains and their corresponding state configurations, which complicates the task of encompassing all potential scenarios through data-driven methodologies alone. Moreover, there are legitimate concerns regarding the generalization and reliability of such approaches, as they often overlook inherent physical constraints. In response to these challenges, this study introduces a novel machine-learning architecture that is highly generalizable and adheres to conservation laws and physical symmetries, thereby ensuring greater reliability. The foundation of this architecture is graph neural networks (GNNs), which are adept at accommodating a variety of shapes and forms. Additionally, we explore the parallels between GNNs and traditional numerical solvers, facilitating a seamless integration of conservative principles and symmetries into machine learning models. Our findings from experiments demonstrate that the model's inclusion of physical laws significantly enhances its generalizability,  i.e., no significant accuracy degradation for unseen spatial domains while other models degrade.
The code is available at \url{https://github.com/yellowshippo/fluxgnn-icml2024}.
\end{abstract}

\section{Introduction}
\label{sec:introduction}

Predicting physical phenomena, often described by partial differential equations (PDEs), has become a focal point in research due to its relevance across various fields such as product design, disaster prevention, and environmental sciences. The integration of machine learning in this domain has been the subject of active investigation recently. This is largely because machine learning has the potential to expedite prediction processes by effectively leveraging existing datasets \cite{ladicky2015data,kochkov2021machine,pichi2024graph}. Moreover, when trained with actual measured data, these models can achieve enhanced prediction accuracy \cite{lam2023learning}. Additionally, such models are increasingly being utilized for optimization and solving inverse problems through gradient-based optimization techniques \cite{chang2020learning,miao2023vc}, facilitated by the efficiency of modern deep learning frameworks in automatic differentiation.

\begin{figure}[tb]
  \centering
  \centerline{\includegraphics[trim={0cm 1cm 22cm 0cm},width=0.9\columnwidth]{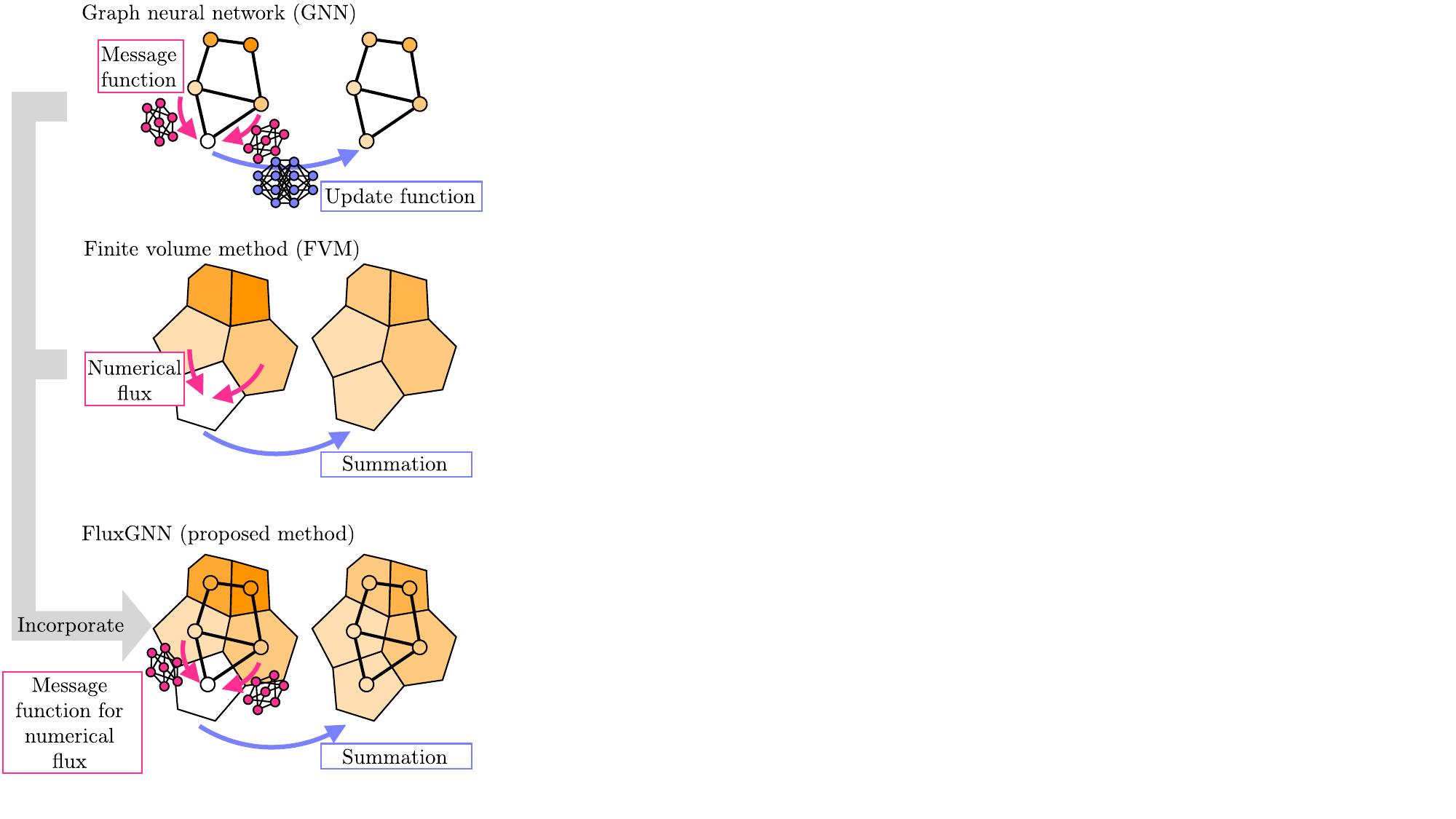}}
  \caption{Overview of a typical graph neural network (GNN), finite volume method (FVM), and proposed model, FluxGNN\@. Our method combines GNN and FVM, realizing high expressibility from GNN and generalizability from FVM.}
  \label{fig:fluxgnn}
  \vskip -0.2in
\end{figure}
Despite these advancements, challenges in generalizability and reliability persist in machine learning applications within physics. The diversity of spatial analysis domains and the myriad of potential states within these domains pose significant difficulties in encompassing all scenarios using solely data-driven approaches. Out-of-domain generalization is challenging in machine learning, leading to reduced generalizability in data-driven models \cite{bonfanti2023hyperparameters}. These models often struggle as surrogate models for classical solvers. Moreover, machine learning methods frequently contravene critical physical laws such as conservation and symmetry, resulting in problematic decision-making processes and less reliable predictions. For instance, in predicting the movement of greenhouse gases on Earth, it is imperative to account for the conservation law of gases, which dictates that the total amount of gas should remain constant barring any generation or absorption.

Incorporating physical laws into machine learning models is a potent strategy to overcome challenges related to generalizability and reliability \cite{ling2016reynolds,wang2021incorporating,huang2022equivariant,li2023rapidly}. Physical systems exhibit symmetries; they remain invariant under transformations such as rotation and scaling. Leveraging these symmetries can enhance a model's generalizability, as an unseen input may be regarded as a familiar one when transformed. Moreover, embedding physical laws as inductive biases in a model can improve the reliability of its predictions, ensuring adherence to these laws.

In this study, we introduce the concept of flux-passing graph neural networks (\emph{FluxGNNs}), which are designed to integrate conservation laws and physical symmetries pertaining to similarity transformations---including translation, rotation, reflection, and scaling---as inductive biases. This approach is grounded in the parallels between graph neural networks (GNNs), which are adaptable to arbitrary graphs, and the finite volume method (FVM), a classical numerical analysis method that inherently respects conservation and symmetries (\cref{fig:fluxgnn}). This integration results in a model that not only generalizes effectively across spatial domains but also rigorously upholds conservation laws and symmetries regarding similarity transformations. The FluxGNNs represent a natural extension of the FVM, essentially encompassing a pure FVM framework enhanced with automatic differentiation capabilities, thanks to its implementation in a deep learning framework. Additionally, we delineate the mathematical conditions for typical GNNs to serve as conservative models.

Our method is designed to seamlessly integrate with existing machine learning techniques. It can accommodate methods such as scaling invariance \cite{wang2021incorporating}, temporal bundling \cite{brandstetter2022message}, and boundary encoder and neural nonlinear solver \cite{horie2022physics}, requiring minimal or no modifications to ensure adherence to conservation laws and symmetries. Additionally, our approach capitalizes on the geometric attributes of the input discretized shapes, such as meshes, by utilizing elements such as surface areas and normal vectors. This integration strengthens the connection between scaling symmetry and fundamental concepts of numerical analysis such as the Courant number. Our experimental results indicate that the proposed method shows high generalizability and improves the speed--accuracy tradeoff of prediction.

The main contributions of our study are summarized below.
\begin{itemize}
  \item
    We mathematically reveal the conditions for typical GNNs to achieve conservations, resulting in the construction of locally conservative GNNs that can be applied to arbitrary graphs.
  \item
    We construct FluxGNNs based on the locally conservative GNNs. The proposed method can be applied to arbitrary analysis domains and satisfy conservation laws rigorously, in addition to achieving similarity-equivariance while focusing on a close relationship with a classical numerical analysis method.
  \item
    We provide methods to incorporate existing machine learning approaches into our model and demonstrate that they can satisfy conservation laws and symmetries under our modifications, if any.
  \item
    Finally, we demonstrate that the proposed method shows high generalizability for unseen spatial domains because of the included physical laws.
\end{itemize}

\section{Background and Related Work}
\label{sec:background}

\subsection{Classical PDE Solvers and FVM}
We focus on a class of PDEs called conservation form in a $n$-dimensional domain $\Omega \in \R^n$ expressed as
\begin{align}
  \pdiff{}{t} \vu = - \nabla \cdot \mF(\vu) + \mS,
  \label{eq:conservation_form}
\end{align}
where
$\mathcal{U} \ni \vu: [0, \infty) \times \Omega \to \R^d$,
$\mF: \mathcal{U} \to \mathcal{U}$, and $\mS$ represent an unknown $d$-dimensional vector field to be solved, a known operator that may be nonlinear and contain spatial differential operators, and a known source term, respectively. This formulation indicates that
$\vu$ is conservative if $\mS = \vzero$ \cite{leveque1992numerical}, and therefore, it is called the conservation form. Although this form looks specific, various PDEs such as the convection--diffusion equation and Navier--Stokes equations can be expressed as conservation forms.
For example, in the case of the diffusion equation with diffusion coefficient $D$, $\mF_\mathrm{diffusion}(u):= D \nabla u$, where $u$ represents a scalar field ($d = 1$ in this case).

FVM\footnote{Although many variants exist for FVM, we introduce a fundamental formulation. For more details on FVM, refer to, e.g., \citet{jasak1996error,darwish2016finite}.} spatially discretizes the equation using meshes, which are discretized representations of the analysis domains. By applying spatial integration over the $i$-th cell (with volume $V_i$) in a mesh and Stokes' theorem, \cref{eq:conservation_form} becomes
\begin{align}
  \pdiff{}{t} \int_{V_i} \vu dV = - \int_{\partial V_i} \mF(\vu) \cdot \vn dS + \int_\Omega \mS dV,
  \label{eq:integral_conservation_form}
\end{align}
where
$\partial V_i$ and $\vn: \partial V_i \to \R^n$ represent the boundary of the $i$-th cell and the unit normal vector field pointing outside of the cell, respectively.

Using linear interpolation, \cref{eq:integral_conservation_form} can be approximated as
\begin{align}
  \pdiff{}{t} V_i \vu_i = - \sum_{j \in \gN_i} S_{ij} \vn_{ij} \cdot \left[ \mF(\vu) \right]_{ij} + \mS_i V_i,
  \label{eq:fvm}
\end{align}
where
$\gN_i$, $\vu_i$, $\mS_i$, $S_{ij}$, $\vn_{ij}$, and $[\mF(\vu)]_{ij}$ represent the set of neighboring cells to the $i$-th cell, value of $\vu$ at the centroid of the $i$-th cell, value of $\mS$ at the centroid of the $i$-th cell, area of the face $(i, j)$, normal vector of the face $(i, j)$ pointing outside of the $i$-th cell, and value of $\mF(\vu)$ at the face $(i, j)$, referred to as \emph{numerical flux}, respectively. Extant studies have developed various computational methods for the numerical flux because the values at the faces need to be estimated using the ones at the cell centers. The conservation law is expressed as $\int_\Omega \vu dV \approx \sum_i V_i \vu_i = \mathrm{Constant}$.

FVM holds local conservation, which is a significant property for solving conservation forms, i.e., the balance between the influx from the $j$-th cell to the $i$-th cell neighboring through the face $(i, j)$, $\bm{\psi}_{ij}:= S_{ij} \vn_{ij} \cdot [\mF(\vu)]_{ij}$, and the opposite, $\psi_{ji}$, expressed as
\begin{align}
  \bm{\psi}_{ji} = S_{ji} \vn_{ji} \cdot [\mF(\vu)]_{ji} = - S_{ij} \vn_{ij} \cdot [\mF(\vu)]_{ij} = - \bm{\psi}_{ij},
  \label{eq:fvm_local_conservation}
\end{align}
using the relationships $S_{ij} = S_{ji}$, $[\mF(\vu)]_{ij} = [\mF(\vu)]_{ji}$, and $\vn_{ij} = - \vn_{ji}$.
Because $\bm{\psi}_{ij} + \bm{\psi}_{ji} = \vzero$, interchanging the physical quantity $\vu$ through face $(i, j)$ does not change the total integral of $\vu$, and therefore, the global conservation laws hold if $\mS_i = \vzero$ ($\forall i \in \gV$). This property inspired us to develop the proposed method that has local conservation.

\subsection{Machine Learning Models for Conservation Laws}
Driven by the critical need for conservation, extensive research has focused on exploring approaches to realize conservation using machine learning. For example, \citet{matsubara2020deep} included structure-preserving numerical methods into their models to obtain conservations in Hamiltonian systems. Further, \citet{richter2022neural} leveraged physics-informed neural networks to parametrize the vector potential, thereby generating a divergence-free, i.e., conservative, vector field. \citet{hansend2024133952} focused on FVM and demonstrated its effectiveness in one-dimensional datasets. These approaches provided insights to achieve conservation and focused on regular or fixed domains; however, they did not investigate irregular domains and spatial extrapolation, which are the main focuses of this study.

\subsection{Machine Learning Models for Symmetries}
Equivariance can be incorporated into machine learning models to obtain models that respect symmetries. Equivariance for a group $G$
is expressed as $f: X \to Y$ satisfying $f(g \cdot x) = g \cdot f(x)$ for any $g \in G$ and $x \in X$, assuming that $G$ acts on $X$ and $Y$. Equivariance for $\En$, i.e., translation, rotation, and reflection, or $SE(n)$, i.e., $E(n)$ without refletion, has been investigated in various works \cite{thomas2018tensor,fuchs2020se,satorras2021n,eijkelboom2023n} and found to be effective in machine learning tasks related to geometry and physics. For example, \citet{horie2021isometric} realized efficient $\En$-equivariant GNNs by using a message-passing method inspired by a classical PDE solver and applying simple $\En$-equivariant multilayer perceptrons (MLPs) expressed as
$\vf_\En(\vh):= \mathrm{MLP}(\Vert \vh \Vert) \vh$,
where $\vh$ and $\Vert \vh \Vert$ represent an input to the layer and its $L^2$-norm, respectively. Further, scale equivariance has been examined extensively in several studies \cite{worrall2019deep,Sosnovik2020Scale-Equivariant,yang2023scale}.
For example, \citet{wang2021incorporating} used a simple formulation of scale-equivariant convolutional neural networks (CNNs) expressed as
$\vf_\mathrm{scaling}(\vh):= \mathrm{CNN}(\vh / \sigma) \sigma$,
where $\sigma:= \max [\vh] - \min [\vh]$ represents the scale of $\vh$.
Further, \citet{wang2021incorporating} demonstrated that $\En$- and scale- equivariance separately improved the predictive performance of machine learning models on fluid dynamics datasets. However, they did not construct a similarity-equivariant model, i.e., an equivariant model considering both $\En$ and scaling. We successfully constructed and demonstrated the effectiveness of such a model in this study.

\subsection{GNNs and Graph Neural PDE Solvers}
A graph $\gG = (\gV, \gE)$ is defined as a tuple of the vertex set $\gV \subset \Z$ and edge set $\gE \subset \Z \times \Z$. GNNs should be able to handle arbitrary-length data because different graphs can have a different number of vertices and different edge connectivities.
Consider a vertex signal $\mH: \gV \to \R^d$ that corresponds to vectors on the vertices of a graph. For a graph with three vertices,
the vertex signal can be expressed as
$\mH:= (\vh_1, \vh_2, \vh_3)^\top$ where $\vh_i \in \R^d$ represents a signal at the $i$-th vertex.
A GNN $\gF: \gH \to \gH'$ can be considered an operator converting a vertex signal to another, where $\gH$ and $\gH'$ denote sets of vertex signals. A message-passing neural network (MPNN) \cite{gilmer2017neural}, which is a typical formulation of GNNs, is expressed as
\begin{equation}
  \begin{aligned}
    \vm_{ij} &= \vf_\mathrm{message}(\vh_i, \vh_j, \vh_{ij}),
    \\
    \vh_i^\mathrm{out} &= \vf_\mathrm{update} \left(\vh_i, \sum_{j \in \gN_i} \vm_{ij} \right),
    \label{eq:mpnn}
  \end{aligned}
\end{equation}
where $\vh_{ij}$ and $\gN_i$ represent a signal on the edge $(i, j)$ and a set of vertices neighboring the $i$-th vertex, respectively. MPNN comprises two neural networks: a message function $\vf_\mathrm{message}$ expressing the interaction between vertices and an update function $\vf_\mathrm{update}$ describing the vertex-wise updates of features.

In the context of spatiotemporal discretization, a typical neural PDE solver is formulated as $\gF: \gH \ni \mH^t \mapsto \mH^{t+1} \in \gH$, where its objective to predict the time series of the vertex signal $\mH^t$.
Given that practical physical simulations often involve irregular spatial data, such as point clouds and meshes, which can naturally be represented as graphs, GNNs emerge as a natural choice for constructing neural PDE solvers \cite{chang2020learning, sanchez2020learning, pfaff2021learning, lam2023learning}.

Notably, \citet{brandstetter2022message} have proposed effective methods for learning and predicting time-dependent PDEs, drawing inspiration from classical PDE solvers. Among their techniques, temporal bundling stands out, as it involves predicting states at multiple time steps through a single forward computation of a model followed by the application of 1D CNN layers in the temporal direction. This approach results in significantly enhanced computational efficiency.

Furthermore, \citet{horie2022physics} have developed PDE solvers based on $\En$-equivariant GNNs, as initially proposed by \citet{horie2021isometric}. Their work is characterized by a rigorous treatment of boundary conditions and an efficient time evolution computation method inspired by nonlinear optimization techniques. They introduced boundary encoders, a concept in which encoders for boundary conditions are combined with those for the corresponding input vertex signals. This allows for the encoding of boundary conditions within the same space as vertex signals. The proposed time evolution method is expressed as
\begin{align}
  \mH\sups{i+1}:=
  \mH\sups{i} - \alpha_\mathrm{BB}\sups{i} \left[
    \mH\sups{i} - \mH\sups{0} - \gF(\mH\sups{i}) \Delta t
    \right],
  \label{eq:neural_nonlinear_solver}
\end{align}
\vskip -10pt
where $\cdot\sups{i}$, $\mH\sups{0} = \mH(t)$, and $\alpha_\mathrm{BB}\sups{i}$ represent the state at the $i$-th nonlinear optimization step, a known graph signal at time $t$, and a coefficient determined using the Barzilai--Borwein method \cite{barzilai1988two}, which contains global pooling operations, respectively. Most existing graph neural PDE solvers use the encode-process-decode architecture proposed by \citet{battaglia2018relational}, wherein input features are embedded into a higher dimensional space using the encoder, message passing is computed in the encoded space, and encoded features are converted into an output space using the decoder, as done in this study.

\section{Proposed Method}
\label{sec:method}
Here, we introduce the proposed method, FluxGNN\@. We first describe an abstract and general formulation of locally conservative GNN, then build FluxGNN on top of that, incorporating the essence of FVM\@. Proofs of all theorems and lemmas claimed in this section can be found in \cref{app:proof}.

\subsection{Locally Conservative GNN}
In this study, we focus on a set of undirected graphs, i.e., $(i, j) \in \gE$ implies $(j, i) \in \gE$ for any $i, j \in \gV$. This framework is employed to represent the exchange of physical quantities between vertices. The condition for a conservative MPNN is outlined as follows:
\begin{theorem}
\label{thm:locally_conservative_gnn}
  An MPNN $\gF: \gH \to \gH$, formulated in \cref{eq:mpnn} with continuous activation functions, exhibits conservation properties for any graphs and vertex signals (i.e., $\sum_{i \in \gV} \vh_i$ remains constant for a given graph) if and only if the following conditions are satisfied.
  \begin{align}
    \vm_{ij} &= - \vm_{ji}, \ \ \forall (i, j) \in \gE
  \label{eq:locally_conservative_message}
    \\
    \vf_\mathrm{update} \left(\vh_i, \sum_{i \in \gN_i} \vm_{ij} \right) &= \vh_i + \sum_{i \in \gN_i} \vm_{ij}.
  \label{eq:locally_conservative_update}
  \end{align}
\end{theorem}
We categorize GNNs that meet the criteria outlined in \cref{thm:locally_conservative_gnn} as \emph{locally conservative GNNs}. The manner in which information is exchanged in these GNNs bears resemblance to the concept of numerical flux in the FVM, specifically concerning the preservation of local conservation principles, as illustrated in \cref{eq:fvm_local_conservation}. The proof of this theorem necessitates the introduction of the following lemma.
\begin{lemma}
 A vertex-wise continuous update function should be linear to achieve conservation.\footnote{One can construct a conservative continuous nonlinear function despite the lemma, unless vertex-wise. For example, the following $\bar{\gF}$ is conservative.
  $\bar{\gF}(\mH):= \gF(\mH) [\sum_{i \in \gV} \mH / \sum_{i \in \gV} \mathcal{F}(\mH)]$.
  However, these are not locally conservative in general, and therefore, we omit these nonlinearities for generalization in this study.
  }
  \label{lem:linear}
\end{lemma}
Following this lemma, the condition for the conservative encode-process-decode architecture can be obtained as
\begin{theorem}
  An encode-process-decode architecture in the form
  $\gF_\mathrm{e\hbox{-}p\hbox{-}d} := \gF_\mathrm{decode} \circ \gF_\mathrm{L}^{(N)} \circ \dots \circ \gF_\mathrm{L}^{(1)} \circ \gF_\mathrm{encode}$
  with locally conservative GNNs $\gF_\mathrm{L}^{(i)}$, a vertex-wise continuous encoder $\gF_\mathrm{encode}$, and a decoder $\gF_\mathrm{decode}$
  is conservative if and only if the encoder is linear and the decoder is the pseudoinverse (more specifically, left inverse) of the encoder, i.e., $\gF_\mathrm{decode} \circ \gF_\mathrm{encode} = \mathrm{id}$, where $\mathrm{id}$ is the identity map.
  \label{thm:encode_process_decode}
\end{theorem}
Interestingly, while \citet{horie2022physics} proposed the use of pseudoinverse decoders to satisfy boundary conditions, our approach adopts them primarily to ensure conservation. Nevertheless, we can also utilize pseudoinverse decoders to fulfill boundary conditions in our context.
In summary, locally conservative GNNs are characterized by their nonlinearity being confined to the message passing functions, without involving update ones.

\subsection{FluxGNN}
Based on the formulation of the FVM (\cref{eq:fvm}) and the locally conservative GNN (\cref{eq:locally_conservative_message,eq:locally_conservative_update}), we construct a class of GNN layers $\gF_\mathrm{Flux}$ expressed as
\begin{align}
  V_i [\gF_\mathrm{Flux}(\mH)]_i:= V_i \vh_i - \sum_{j \in \gN_i} S_{ij} \vn_{ij} \cdot \mF_\mathrm{ML} \left(\vh_i, \vh_{j}, \vh_{ij} \right),
  \label{eq:fluxgnn}
\end{align}
where $\mF_\mathrm{ML}$ and $\vh_{ij}$ represent a machine learning model constructed using neural networks and a feature at the face $(i, j)$ corresponding to an edge feature in MPNN, respectively\@. Further, vertices correspond to the cells of the mesh, as shown in \cref{fig:fluxgnn}. We refer to $\mF_\mathrm{ML}$ as a \emph{flux function} because it is a message function for the numerical flux.
\begin{theorem}
  A machine learning model defined in \cref{eq:fluxgnn} with a symmetric edge signal $\vh_{ij} = \vh_{ji}$ is a locally conservative GNN if and only if
  a flux function $\mF_\mathrm{ML}$ is a permutation-invariant function for the first and second arguments, i.e.,
  $\mF_\mathrm{ML}(\vh_i, \vh_j, \vh_{ij}) = \mF_\mathrm{ML}(\vh_j, \vh_i, \vh_{ij}) %
  $
  \label{thm:fluxgnn}
\end{theorem}
We define models $\gF_\mathrm{Flux}$ described in \cref{eq:fluxgnn} as \emph{FluxGNNs}, when the condition outlined in \cref{thm:fluxgnn} is satisfied, with encoders and decoders also adhering to the criteria specified in \cref{thm:encode_process_decode}. FluxGNNs are designed to model numerical fluxes using machine learning techniques, hence the name. The attainment of the former condition can be realized through the utilization of deep sets, as presented in \cite{zaheer2017deep}, which offer the necessary and sufficient class of permutation-invariant functions. FluxGNN exhibits, at a minimum, the same level of expressive power as the FVM concerning information propagation in space. This equivalence is attributed to the presence of the formulation of spatial discretization in FVM (\cref{eq:fvm}) within \cref{eq:fluxgnn}. Further, FluxGNN demonstrates similarity equivariance when $\mF_\mathrm{ML}$ exhibits such properties. It is because other components, such as area $S_{ij}$, volume $V_i$, and normal vector $\vn_{ij}$, are considered geometrical quantities, and therefore, exhibit similarity-equivariance.

\subsection{Incorporation of Existing Methods}
FluxGNN is designed with a level of simplicity that allows for compatibility with numerous existing methods. In this section, we explore how to integrate certain aspects of these methods while maintaining conservation. All operations are vertex-wise, and therefore, we write $\vh$ instead of $\vh_i$ for the $i$-th vertex feature ($i \in \gV$) for simplicity.

\subsubsection{Similarity-Equivariant MLPs}
Similarity-equivariant MLPs can be achieved by combining the existing $\En$-equivariant layer and scale-equivariant layer as
\begin{align}
  \vf_\mathrm{sim}(\vh):= \mathrm{MLP}(\Vert \vh / \sigma \Vert) \vh,
  \label{eq:similarity_equivariant_mlp}
\end{align}
where $\sigma > 0$ is determined using local quantities, such as spatial resolution $\Delta x$ and temporal resolution $\Delta t$, to make the input physical quantities dimensionless. We avoid using global operators, such as $\min$ and $\max$ used in \citet{wang2021incorporating}, to retain the localness of our model. Further, our choice of $\sigma$ has a strong connection with central concepts in classical PDE solvers. For example, assuming that the input feature is in the dimension of velocity, our similarity equivariant layer turns to
\begin{align}
  \mathrm{MLP}\left(\left\Vert \frac{\vh}{\Delta x / \Delta t} \right\Vert \right) \vh
  =
  \mathrm{MLP}(C) \vh,
\end{align}
where $C$ represents the Courant number that characterizes the stability of the simulation. This formulation is natural because a condition with a different Courant number can require a different treatment for efficient computations because of a different numerical stability. In the case of the diffusion phenomena, the direct input to MLP turns to the diffusion number, which is another essential quantity for numerical stability.
In addition, \citet{huang2022equivariant} suggest a similar form to our formulation, showing its universality. That justifies our choice of the similarity-equivariant MLPs.
Using similarity-equivariant MLPs as building blocks, one can construct flux functions as presented in \cref{app:flux_function}.

\subsubsection{Temporal Bundling}
Due to the constraints imposed by encoders and decoders on conservation, it is necessary to modify the temporal bundling method. In scenarios where we predict $K$ steps simultaneously using a single forward computation of a FluxGNN, the approach involves dividing the weight matrices of the encoder and decoder into $K$ distinct segments. Each of these segments is specifically applied to encode or decode features. Subsequently, we concatenate these processed features in the feature direction, as follows:
$\mathrm{Concat}\left[ \mW\supp{1} \vh, \mW\supp{2} \vh, \dots, \mW\supp{K} \vh \right]$.
Once the decoding process is completed using the pseudoinverse decoder, the next step involves the application of multiple linear 1D-CNN layers. One such layer, exemplifying this approach, can be represented in the following form:
$\vh\supp{t}_\mathrm{out}:= (1 - w) \vh\supp{t - 1} + w \vh\supp{t}$,
where $w$ represents a trainable parameter and $\vh\supp{t}$ denotes the outputs of decoders at time $t$, to facilitate the temporal smoothness. This form differs from that proposed in \citet{brandstetter2022message} because we need this operation to be linear for conservation.

\subsubsection{Neural Nonlinear Solver and Boundary Encoder}
The neural nonlinear solver (\cref{eq:neural_nonlinear_solver}) can be applied to FluxGNN without modification because it does not break conservation laws and symmetries. Although it contains global operations to compute $\alpha_\mathrm{BB}^{[i]}$, the model remains locally conservative because it exists outside of MLPs. Further, boundary encoders do not need modifications because they are encoders, which are linear in our formulation.

\section{Experiments}
\label{sec:experiments}
We showcase the outcomes of our experiments, demonstrating that our proposed method achieves conservation and similarity-equivariance with notable accuracy and computational efficiency. We have implemented all our models using PyTorch 1.9.1 \cite{NEURIPS2019_9015}. The code is available at \url{https://github.com/yellowshippo/fluxgnn-icml2024}. For a comprehensive overview and detailed insights into the experiments, please refer to \cref{app:cd,app:mixture}.

\begin{figure}[t]
  \centering
  \includegraphics[trim={0cm 9cm 0cm 0cm},width=0.99\columnwidth]{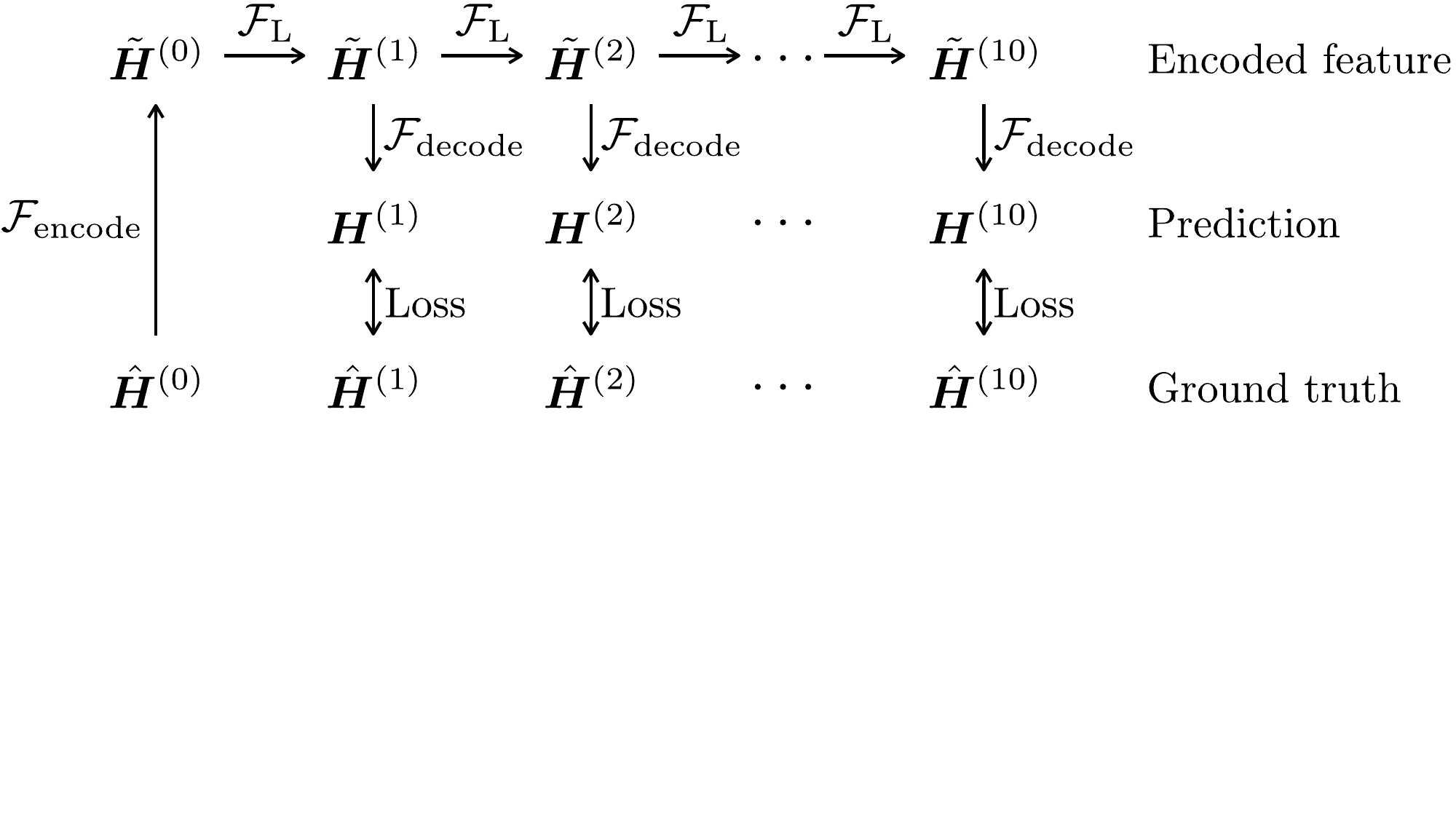}
    \caption{Overview of the FluxGNN model for the convection--diffusion equation with an encoder $\gF_\mathrm{encode}$, locally conservative GNN $\gF_\mathrm{L}$, and decoder $\gF_\mathrm{decode}$. The model outputs a time series, and the loss is computed using all steps of the model's output.}
  \label{fig:model_cd}
  \vskip -0.1in
\end{figure}

\begin{figure}[tb]
  \centering
  \centerline{\includegraphics[trim={0cm 1cm 0 0cm},width=0.99\columnwidth]{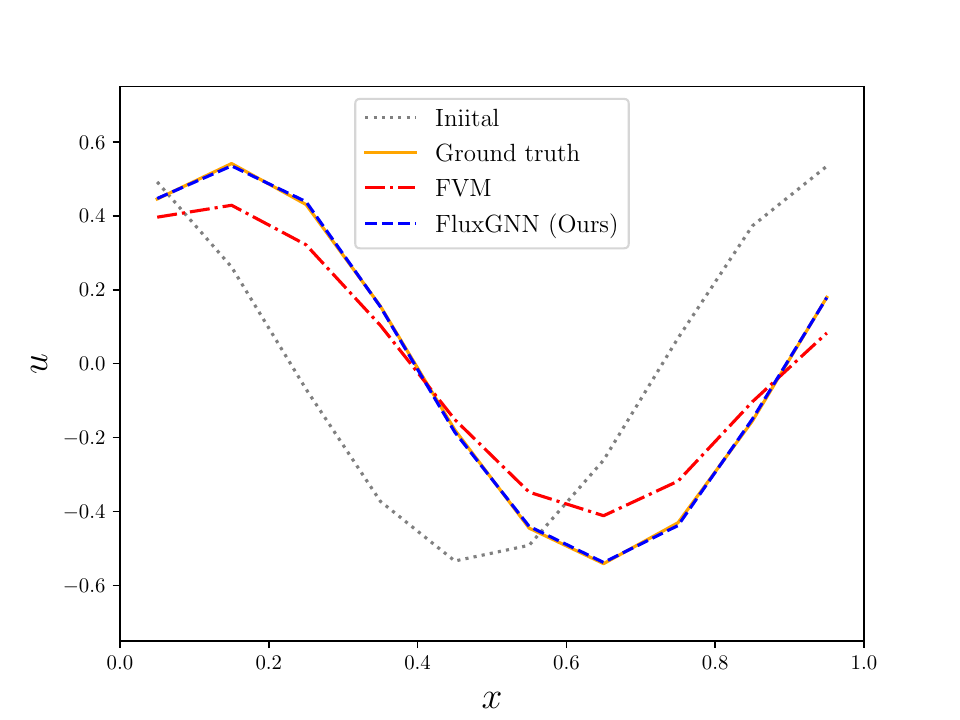}}
    \caption{Comparison of the initial condition, ground truth, prediction of FVM, and prediction of FluxGNN taken from a sample in the test dataset at time $t = 1.0$.}
  \label{fig:cd}
  \vskip -0.1in
\end{figure}

\subsection{Convection--Diffusion Equation}
\label{sec:cd}
In the current set of experiments, we demonstrate the expressive capabilities of the core FluxGNN model using the convection--diffusion equation, a fundamental model for the transport of conservative quantities. The equation for a conservative scalar field $u$ is expressed as
\begin{align}
  \pdiff{}{t} u = - \nabla \cdot \left(\vc u - D \nabla u \right),
\end{align}
where $\vc$ and $D \geq 0$ represent a known velocity field and known diffusion coefficient, respectively.

For our datasets, we generated 100 trajectories for training, 10 for validation, and 10 for testing. These were derived using the exact solution of the equation, with random variations in uniform velocity $u$ from 0.0 to 0.2, and in the amplitude and phase of the sinusoidal initial condition. We set the diffusion coefficient $D = 10^{-4}$, corresponding to convection-dominant problems, which are generally more challenging than diffusive ones. The spatial and temporal resolution parameters were set to $\Delta x = 0.1$ and $\Delta t = 0.1$, respectively, and the analysis domain length was one unit.

We constructed a straightforward FluxGNN model with autoregressive time series modeling, as illustrated in Figure \cref{fig:model_cd}. The model generates time series data from given initial conditions and velocities. It was trained on a CPU (Intel Xeon CPU E5-2695 v2 @ 2.40 GHz) for 3 hours, using MSE loss and an Adam optimizer \cite{kingma2014adam}. As a baseline, we established an FVM solver making all trainable functions identity ones, enabling most implementations to be shared between the solver and FluxGNN\@. This setup highlights the enhancements brought by machine learning. The accuracy of our FVM implementation was verified across multiple problems (\cref{app:validation}).

\cref{fig:cd} and \cref{tab:cd} present the results of the experiments in comparison with the ground truth and the FVM solver. The conservation error is defined by
$\int_{t \in (0, T_\mathrm{max}]} \left[ \int_\Omega (\mH(t) - \mH(t = 0)) dV \right] dt$,
where
$T_\mathrm{max} = 1.0$.
Our model exhibited higher accuracy than the FVM at the same spatiotemporal resolution. Notably, the FVM tends to exhibit increased diffusivity due to numerical diffusion from the low spatial resolution. Moreover, FluxGNN maintained a conservation property almost equivalent to that of the FVM, indicating that the neural networks incorporated in the model do not disrupt the conservation law. Consequently, we conclude that our machine learning model possesses superior expressive power and can leverage existing data to achieve higher accuracy compared to classical methods.

\begin{table}[tb]
  \vskip -0.1in
  \caption{MSE loss ($\pm$ standard error of the mean) on the test dataset of the convection--diffusion equation.}
  \centering
  \label{tab:cd}
  \begin{small}
    \begin{tabular}{lcc}
      \toprule
      Method
      &
      \makecell{%
        Loss
        \\
        $\left(\times 10^{-5} \right)$
      }
      &
      \makecell{%
        Conservation error
        \\
        $\left(\times 10^{-8} \right)$
      }
      \\
      \midrule
      FVM
      &
      1408.47 $\pm$ 32.75
      &
      1.63 $\pm$ 0.19
      \\
      \textbf{FluxGNN} (Ours)
      &
      \textbf{3.08} $\pm$ 0.08
      &
      1.75 $\pm$ 0.32
      \\
      \bottomrule
    \end{tabular}
  \end{small}
  \vskip -0.1in
\end{table}

\begin{figure*}[tb]
  \centering
  \centerline{\includegraphics[trim={0cm 5cm 0cm 0cm},width=0.95\linewidth]{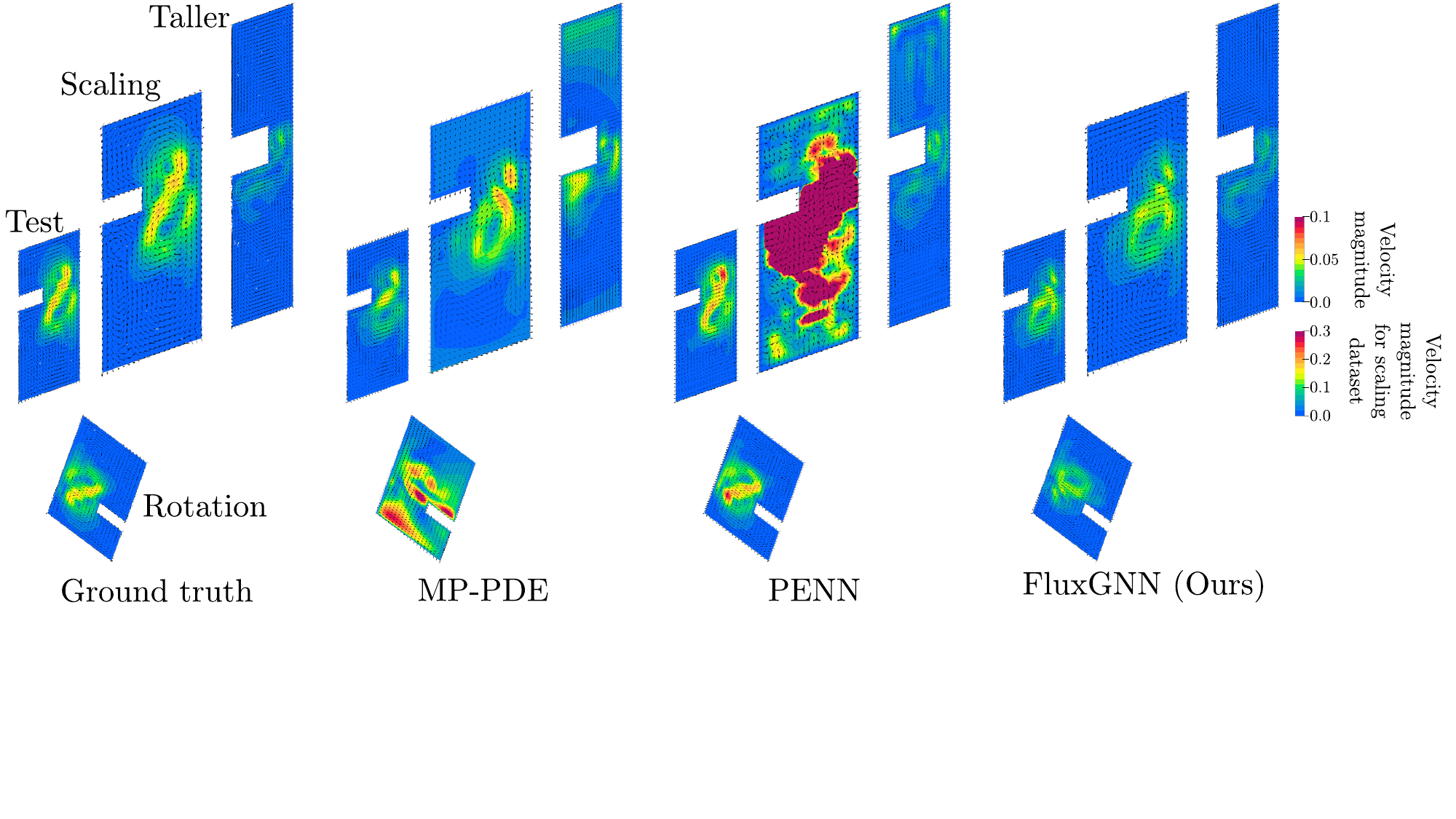}}
  \caption{Visual comparison of the velocity field between the ground truth, MP-PDE, PENN, and FluxGNN.}
  \label{fig:mixture_u}
\end{figure*}

\subsection{Navier--Stokes Equations with Mixture}
\label{sec:mixture}
To evaluate the practical applicability of our model, we conducted tests on a more complex problem, integrating existing methods into FluxGNN\@. This scenario leads to intricate interactions between velocity $\vu$, pressure $p$, and density $\rho$. These interactions are expressed as follows
\begin{align}
  \pdiff{}{t} (\rho \vu)
  =&
  - \nabla \cdot \left[
    \rho \vu \otimes \vu
    - \mu (\nabla \otimes \vu + (\nabla \otimes \vu)^\top)
    \right]
  \nonumber
  \\
  &- \nabla p_{\rho g y} - \vg \cdot \vy \nabla \rho,
  \\
  \pdiff{}{t} \alpha &= - \nabla \cdot \left[
    \vu \alpha - D_\alpha \nabla \alpha
    \right],
  \label{eq:mixture}
\end{align}
where $\mu$ and $\alpha$ represent the dynamic viscosity, volume fraction of the low-density fluid, respectively. $\alpha$ satisfies the relationship
$\rho = \alpha \rho_\mathrm{low} + (1 - \alpha) \rho_\mathrm{high}$
($\rho_\mathrm{low}$ and $\rho_\mathrm{high}$ denote scalar values of low and high density, respectively). Furthermore, $D_\alpha$, $\vg$, $\vy$, and $p_{\rho g y} = p - \rho \vg \cdot \vy$ represent the diffusion coefficient for $\alpha$, gravity acceleration, height in the direction of the gravity, and pressure without the effect of gravitational potential, respectively.

We generated 200 trajectories for training, 8 for validation, and 8 for the test dataset. These trajectories encompassed various shapes and initial conditions, and they were generated using OpenFOAM,\footnote{\url{https://www.openfoam.com/}} a widely recognized classical PDE solver. We generated five different shapes for training, two for validation, and two for testing, with variations in the dimensions characterizing each shape. To ensure realism, we employed random numerical analysis to create the initial conditions for each shape. The primary objective of the machine learning task was to predict ground truth trajectories projected onto domains with lower spatiotemporal resolution. To achieve this, we set the time step $\Delta t = 0.2$ with the maximum time reaching 1.6 for training and 3.2 for evaluation, allowing us to assess the model's ability for temporal extrapolation. In addition to these datasets, we generated two supplementary datasets, each comprising eight trajectories. These datasets were created by applying rotation and scaling operations to the test dataset, respectively (named \emph{rotation} and \emph{scaling} datasets). Further, to test the model's generalizability, we generated a dataset featuring three shapes that were taller than the test samples (named \emph{taller} dataset) to validate the model's generalizability.

We developed FluxGNN models following the autoregressive approach detailed in \cref{sec:cd}, but with the incorporation of components from existing research. These included similarity-equivariant MLPs, temporal bundling, neural nonlinear solvers, and boundary encoders. The model inputs were aligned with those used in classical numerical analysis, including mesh geometry, initial conditions, boundary conditions, and material properties. The model outputs were time series data for velocity ($\vu$), pressure ($p_{\rho gy}$), and volume fraction ($\alpha$) fields. We employed RMSE loss for each physical quantity, weighted by their respective standard deviations, to balance the loss magnitudes across different physical quantities.

For comparison, we selected MP-PDE \cite{brandstetter2022message} and PENN \cite{horie2022physics} as baseline models, representing state-of-the-art machine learning approaches that accommodate irregular domains and boundary conditions.
That comparison highlighted our model's superiority because we incorporated these methods into ours.
Additionally, we included our FVM implementation to examine the enhanced expressibility achieved through machine learning, similar to the approach in \cref{sec:cd}. All machine learning models were trained on GPUs (NVIDIA A100 80GB PCIe) over a period of three days.

\begin{table*}[tb]
  \vskip -0.1in
  \caption{RMSE loss and conservation error ($\pm$ standard error of the mean) on the evaluation datasets of the Navier--Stokes equation with mixture. Each metric is normalized using standard deviation.}
  \centering
  \label{tab:mixture}
  \begin{small}
    \begin{tabular}{lrrrrr}
      \toprule
      Method
      &
      Dataset
      &
      Loss $\vu$ $\left(\times 10^{-1} \right)$
      &
      Loss $p$ $\left(\times 10^{-1} \right)$
      &
      Loss $\alpha$ $\left(\times 10^{-1} \right)$
      &
      Conservation error $\alpha$ $\left(\times 10^{-5} \right)$
      \\
      \midrule
FVM & test &
$19.501 \pm 0.149$ &
$37.706 \pm 0.194$ &
$2.896 \pm 0.030$ &
$\boldsymbol{0.01} \pm 0.00$
\\
MP-PDE & test &
$1.532 \pm 0.010$ &
$0.941 \pm 0.008$ &
$1.021 \pm 0.010$ &
$1001.95 \pm 120.43$
\\
PENN & test &
$\boldsymbol{0.619} \pm 0.005$ &
$\boldsymbol{0.598} \pm 0.005$ &
      $\boldsymbol{0.358} \pm 0.005$ &
$1356.62 \pm 283.00$
\\
FluxGNN (Ours) & test &
$1.202 \pm 0.008$ &
$1.143 \pm 0.008$ &
$\boldsymbol{0.349} \pm 0.005$ &
$0.06 \pm 0.03$
\\[5pt]
FVM & rotation &
$18.363 \pm 0.128$ &
$22.366 \pm 0.123$ &
$2.930 \pm 0.030$ &
$\boldsymbol{0.01} \pm 0.00$
\\
MP-PDE & rotation &
$11.523 \pm 0.075$ &
$10.883 \pm 0.078$ &
$9.528 \pm 0.072$ &
$5307.85 \pm 1009.04$
\\
PENN & rotation &
$\boldsymbol{0.622} \pm 0.004$ &
$\boldsymbol{0.592} \pm 0.005$ &
$\boldsymbol{0.355} \pm 0.005$ &
$1302.98 \pm 153.02$
\\
FluxGNN (Ours) & rotation &
$1.207 \pm 0.007$ &
$1.175 \pm 0.008$ &
$\boldsymbol{0.351} \pm 0.005$ &
$\boldsymbol{0.01} \pm 0.00$
\\[5pt]
FVM & scaling &
$19.411 \pm 0.149$ &
$38.106 \pm 0.193$ &
$2.918 \pm 0.030$ &
$\boldsymbol{0.01} \pm 0.00$
\\
MP-PDE & scaling &
$5.072 \pm 0.034$ &
$4.211 \pm 0.040$ &
$5.604 \pm 0.046$ &
$6401.93 \pm 1806.50$
\\
PENN & scaling &
$\mathrm{NaN} \pm \mathrm{NaN}$ &
$\mathrm{NaN} \pm \mathrm{NaN}$ &
$\mathrm{NaN} \pm \mathrm{NaN}$ &
$\mathrm{NaN} \pm \mathrm{NaN}$
\\
FluxGNN (Ours) & scaling &
$\boldsymbol{1.219} \pm 0.009$ &
$\boldsymbol{1.228} \pm 0.008$ &
$\boldsymbol{0.356} \pm 0.005$ &
$0.05 \pm 0.01$
\\[5pt]
FVM & taller &
$\mathrm{NaN} \pm \mathrm{NaN}$ &
$\mathrm{NaN} \pm \mathrm{NaN}$ &
$\mathrm{NaN} \pm \mathrm{NaN}$ &
$\mathrm{NaN} \pm \mathrm{NaN}$
\\
MP-PDE & taller &
$8.377 \pm 0.081$ &
$2.472 \pm 0.024$ &
$2.823 \pm 0.033$ &
$16770.80 \pm 3293.59$
\\
PENN & taller &
$1.476 \pm 0.016$ &
$1.887 \pm 0.014$ &
$0.357 \pm 0.006$ &
$4799.89 \pm 1574.43$
\\
FluxGNN (Ours) & taller &
$\boldsymbol{1.184} \pm 0.009$ &
$\boldsymbol{0.966} \pm 0.008$ &
$\boldsymbol{0.337} \pm 0.006$ &
$\boldsymbol{0.02} \pm 0.00$
\\
      \bottomrule
    \end{tabular}
  \end{small}
  \vskip -0.1in
\end{table*}

\begin{figure}[bt]
  \centering
  \centerline{\includegraphics[trim={0cm 8.5cm 18cm 0cm},width=0.95\columnwidth]{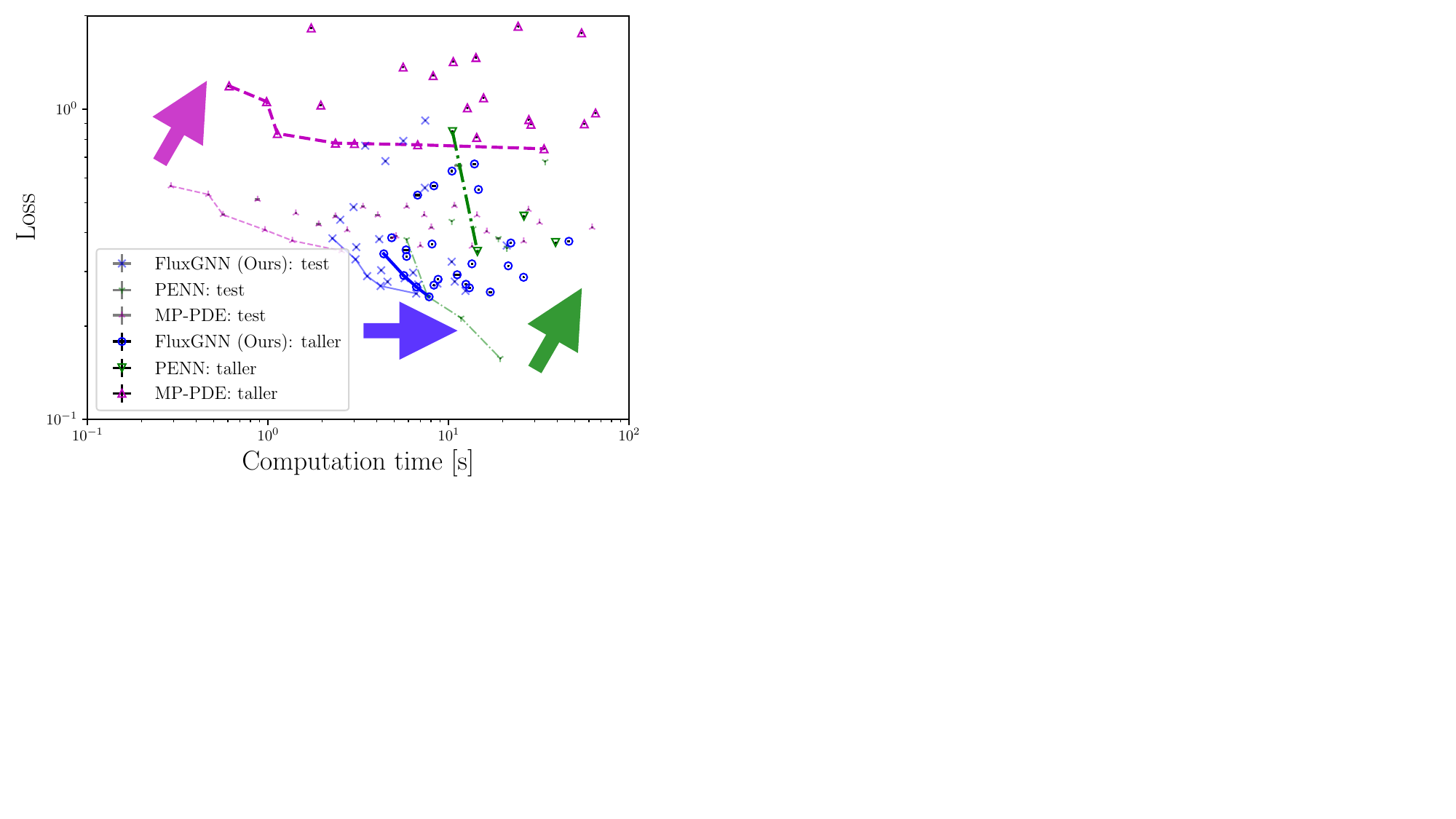}}
    \caption{Speed--accuracy tradeoff obtained through hyperparameter studies for MP-PDE, PENN, and FluxGNN (proposed method), with error bars corresponding to the standard error of the mean. Light and dark colors correspond to the evaluation of the test and taller datasets, respectively. Lines represent Pareto fronts, with arrows denoting shifts of the fronts caused by changes in the dataset considered for the evaluation. The results of FVM are excluded because they are far from the Pareto front and are shown in \cref{fig:mixture_tradeoff_detail}.}
  \label{fig:mixture_tradeoff}
  \vskip -0.2in
\end{figure}

\cref{fig:mixture_u} and \cref{tab:mixture} present both qualitative and quantitative comparisons for each method. All machine learning models demonstrate excellent performance, indicating the success of the training process. Furthermore, all models exhibit the ability to predict time series longer than those observed during training. The proposed method demonstrates a high level of conservation, even when incorporating components from existing machine learning methods. In contrast, other baseline machine learning models do not exhibit the same level of conservation, highlighting the need for special treatment to adhere to conservation laws. When considering the rotation dataset, PENN performs admirably due to its $\En$-equivariance. However, the MP-PDE model begins to degrade under these conditions. On the other hand, PENN displays divergent behavior when applied to the scaling dataset, primarily because of the significant changes in input values, leading to extrapolation. In contrast, FluxGNN exhibits nearly identical performance before and after scaling, showcasing its scale equivariance. Nevertheless, we observe a slight performance variation attributed to numerical errors accumulated during autoregressive computation.

Our method performs well on the taller dataset despite no scaling, which implies that conservation is essential for realizing generalizability because PENN degrades its performance even with $\En$-equivariance and shows the spurious emergence of velocity on the top of the taller sample. Further, our model leverages the expressive power of neural networks because FVM tends to be divergent and never reaches convergence on the taller dataset for all tested settings. Therefore, we claim that FluxGNN incorporates methods of the classical PDE solver and machine learning at a high level, realizing high generalizability and expressibility.

\cref{fig:mixture_tradeoff} presents the speed--accuracy tradeoff obtained using machine learning models with various hyperparameters on one core of the same CPU as \cref{sec:cd}. All models have no clear advantage on the test dataset. However, FluxGNN has a clear advantage compared to the baseline models when evaluated on the taller dataset. Moreover, the Pareto front of our model does not shift significantly in the direction of the loss, which means that FluxGNN generalizes for spatial extrapolation without losing accuracy.

\begin{figure}[tb]
  \centering
  \centerline{\includegraphics[trim={0cm 0cm 0cm 0cm},width=0.7\columnwidth]{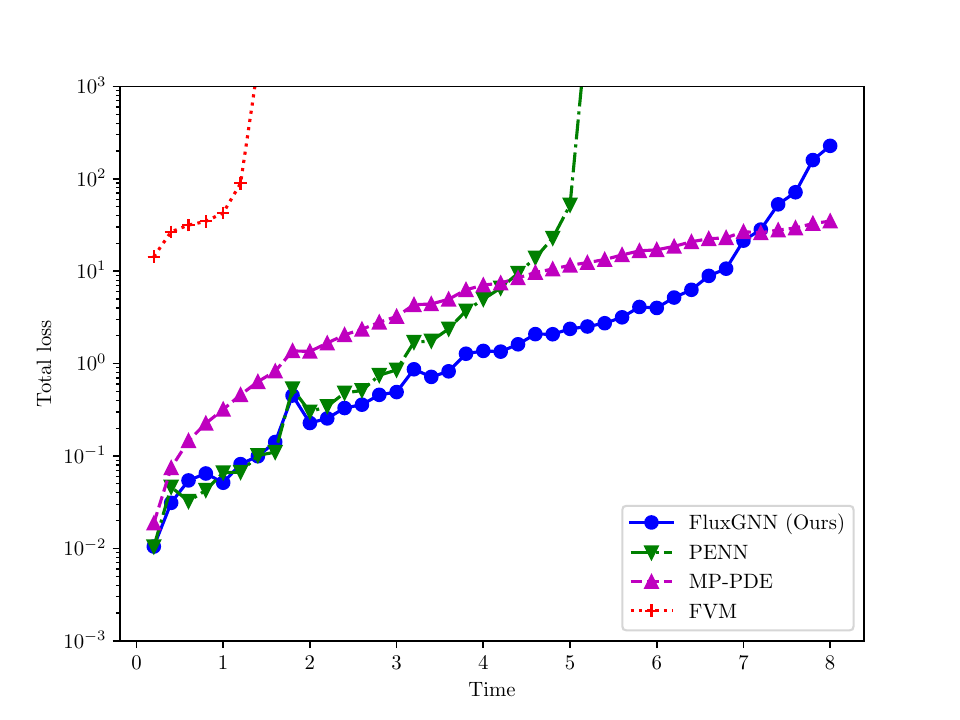}}
    \caption{Time evolution of total loss for FVM, MP-PDE, PENN, and the proposed method, FluxGNN. The vertical axis is in log scale.}
  \label{fig:ts_loss}
\end{figure}

\begin{figure}[bt]
  \centering
  \centerline{\includegraphics[trim={0cm 0cm 0cm 0cm},width=0.7\columnwidth]{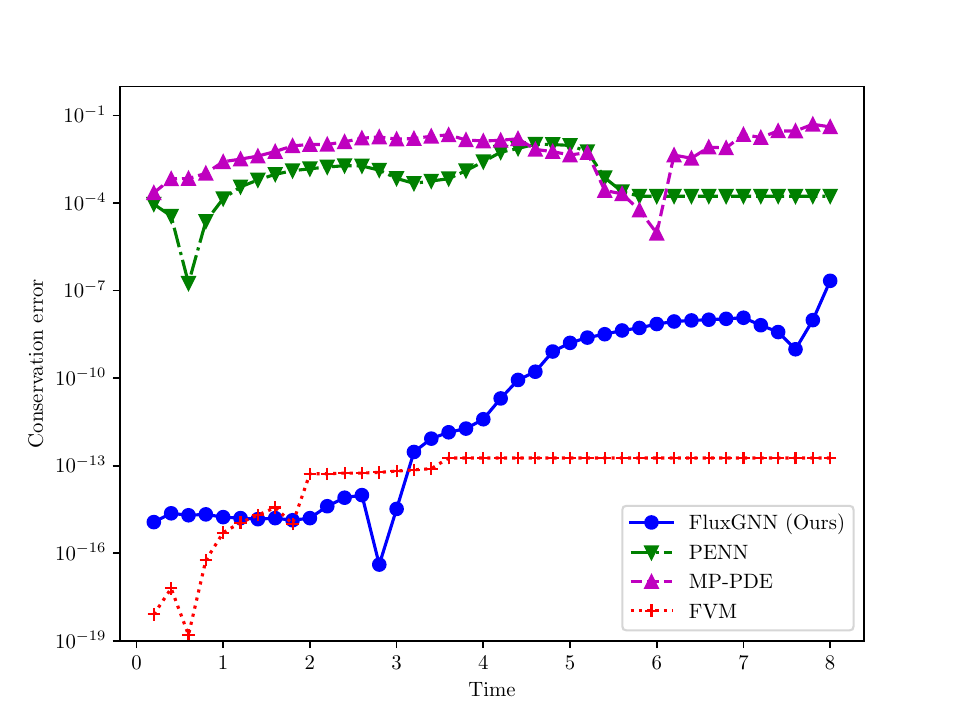}}
    \caption{Time evolution of conservation error for FVM, MP-PDE, PENN, and the proposed method, FluxGNN. The vertical axis is in log scale.}
  \label{fig:ts_cons}
\end{figure}
We performed prediction using each method for even further temporal rollout. \cref{fig:ts_loss} presents the time evolution of total loss. FluxGNN exhibits the lowest loss until time $t \approx 7.0$. After that, it starts to show a higher loss than MP-PDE due to numerical instability. \cref{fig:ts_cons} indicates that the proposed method demonstrates a high level of conservation. Even after instability stands out, the conservation error of our method is lower than other machine learning methods by orders of magnitude. The results imply the potential of FluxGNN for a more extended temporal rollout.
We did not include the pushforward trick proposed by \citet{brandstetter2022message}, which realizes a more stable prediction for time series prediction. Incorporating that trick into our method may address this point. In addition, incorporating flux and slope limiters, known as effective methods for stability in a computational physics domain, may also help.
The visualization of longer rollout predictions is shown in \cref{app:rollout}.

\cref{app:ablation} elaborates on an ablation study of the FluxGNN model. That suggests that all components included in the model contribute to spatial out-of-domain generalizability.

\section{Conclusion}
\label{sec:conclusion}
We presented FluxGNNs, which combines the expressibility of GNNs and the generalizability of FVM. Our method shows high generalizability for spatial domain extrapolation, where other machine learning models exhibit performance degradation. This study focused on PDEs corresponding to flow and transport phenomena; however, it can be applied to a broader class of problems, e.g., optimal transport and traffic flow on graphs because the locally conservative GNNs proposed in this study have a generic structure.

Although our model shows high expressibility and generalizability, we did not demonstrate performance analysis comparing well-optimized classical solvers, which is a limitation of the work. However, we demonstrated a clear improvement compared to FVM, which shared most of the implementation with our model. Therefore, one possible next step would be implementing FluxGNN on top of the well-optimized solvers, which may drastically improve speed and accuracy, as demonstrated in the experiments.

Another limitation is that the method needs to be more stable for long-term prediction. This point may be addressed by incorporating flux and slope limiters, known as effective methods for stability in a computational physics domain. We leave it as a future work because our focus of the present work is to construct a reliable method towards spatially out-of-domain generalization.

\section*{Acknowledgment}
This work was supported by
JST PRESTO Grant Number JPMJPR21O9,
JST FOREST Grant Number JPMJFR215S,
JSPS KAKENHI Grant Numbers 23H04532, 23K24857, and 23KK0182,
and
ATLA ``Innovative Science and Technology Initiative for Security'' Grant Number JPJ004596.
The authors gratefully acknowledge the reviewers of the conference for their helpful comments and fruitful discussions.
We would like to thank Editage for editing and reviewing this manuscript for English language.

\section*{Impact Statement}
This paper presents work whose goal is to advance the field of Machine Learning. There are many potential societal consequences of our work, none which we feel must be specifically highlighted here.

\bibliography{fluxgnn}
\bibliographystyle{icml2024}

\newpage
\appendix
\onecolumn

\section{Notation}

\bgroup
\def\arraystretch{1.5}
\begin{tabular}{p{0.2\textwidth}p{0.7\textwidth}}
  $\displaystyle \Omega$ & Analysis domain
  \\
  $\displaystyle \gU: [0, \infty) \times \Omega \to \R^d$ & Time-dependent vector field
  \\
  $\displaystyle V_i$ & Volume of the $i$-th cell
  \\
  $\displaystyle S_{ij} = S_{ji}$ & Area of the face $(i, j)$
  \\
  $\displaystyle \vn_{ij} = - \vn_{ji}$ & Normal vector of the face $(i, j)$, pointing outside of the $i$-th cell
  \\
  $\displaystyle \vx_i$ & Position of the centroid of the $i$-th cell
  \\
  $\displaystyle \vx_{ij} = \vx_{ji}$ & Position of the centroid of the face $(i, j)$
  \\
  $\displaystyle \V \subset \Z$ & Vertex set
  \\
  $\displaystyle \E \subset \Z \times \Z$ & Edge set
  \\
  $\displaystyle \G = (\V, \E)$ & Graph
  \\
  $\displaystyle \gN_i \subset \gV$ & Set of neighboring vertices regarding the $i$ vertex
  \\
  $\displaystyle \gH$ & Set of graph signals
  \\
  $\displaystyle \mH: \gV \to \R^d$ & Graph signal
  \\
  $\displaystyle \vh_i:= \left[ \mH \right]_i$ & Value of a graph signal at the $i$-th vertex
  \\
  $\displaystyle \gF: \gH \to \gH'$ & Operator mapping from a vertex signal to another, typically constructed using GNNs
  \\
\end{tabular}
\egroup

\section{Basics on Finite Volume Method}
\label{app:fvm}
We review how to model the linear convection--diffusion equation of a scalar field expressed as
\begin{align}
  \pdiff{}{t} u = - \nabla \cdot \left(\vc u - D \nabla u \right),
  \label{eq:cd}
\end{align}
because it is easy to generalize the scheme to vector or higher-order tensor fields.

\begin{figure}[bt]
  \centering
  \centerline{\includegraphics[trim={0cm 0cm 0cm 0cm},width=0.8\linewidth]{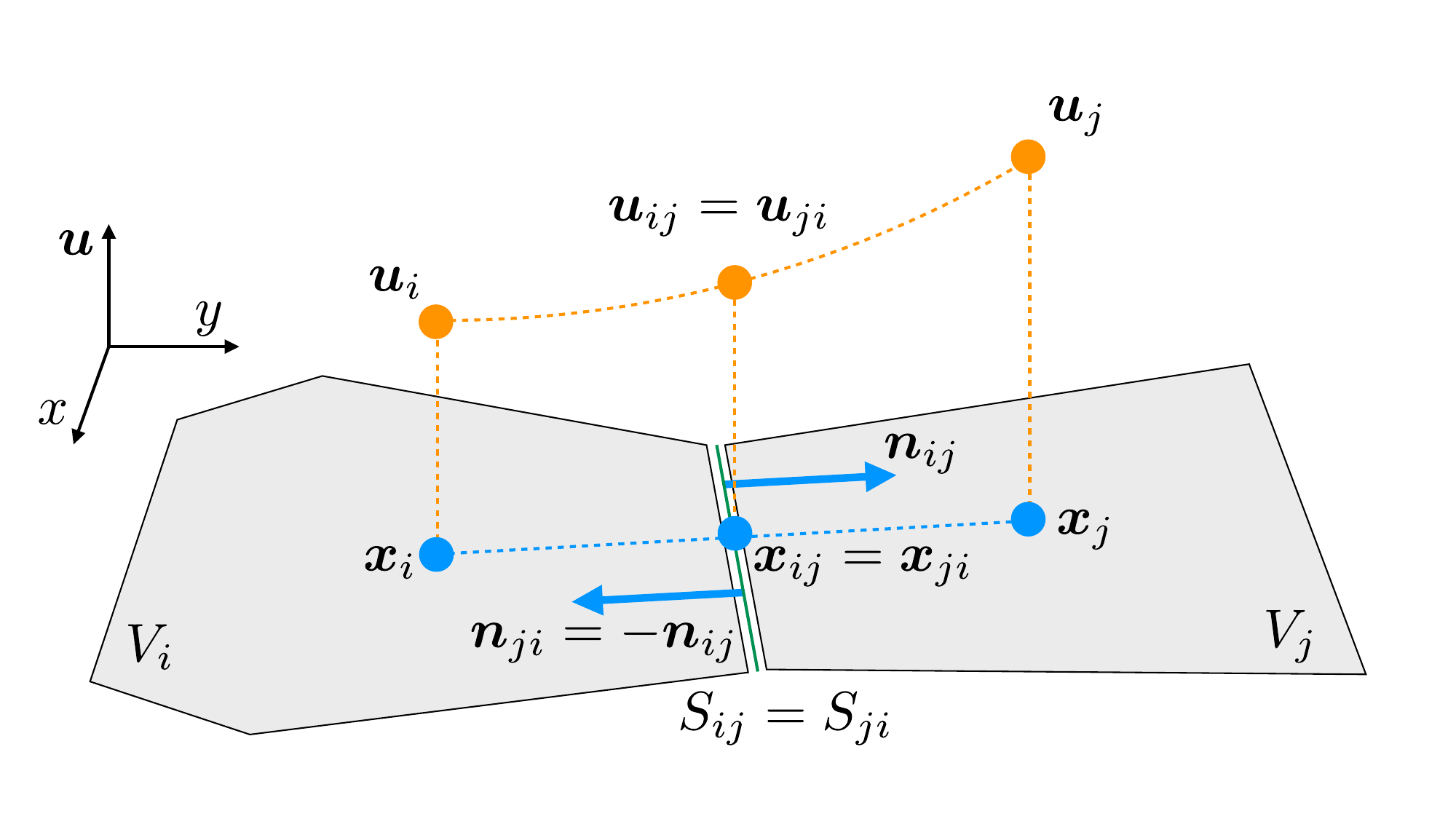}}
    \caption{Geometry and variables used to construct FVM, focusing on the $i$-th and $j$-th cells.
  }
  \label{fig:fvm_setting}
\end{figure}

Here, we consider two neighboring cells, $i$ and $j$, assuming $\vn_{ij}$ and $\vx_i - \vx_j$ are in the same direction, and the midpoint is at the center of the face $(i, j)$, i.e., $(\vx_i + \vx_j) / 2 = \vx_{ij}$, for simplicity.
The setting is illustrated in \cref{fig:fvm_setting}.
In case the assumption does not hold, refer to \citet{jasak1996error,versteeg1995computational,darwish2016finite}.

\subsection{Spatial discretization}
Applying \cref{eq:fvm} to \cref{eq:cd}, we obtain
\begin{align}
  \pdiff{}{t} V_i u_i
  &=
  \sum_{j \in \gN_i} S_{ij} \vn_{ij} \cdot \left[
    \vc u - D \nabla u
    \right]_{ij}
  \\
  &\approx
  \sum_{j \in \gN_i} S_{ij} \vn_{ij} \cdot \left[
    \vc_{ij} u_{ij} - D_{ij} [\nabla u]_{ij}
    \right],
\end{align}
where $\cdot_{ij}$ represents a value at the face $(i, j)$.

To obtain $\vu_{ij}$ for the convection term, one may use the central interpolation
\begin{align}
  u_{ij}^\mathrm{linear}:=
  \frac{\Vert \vx_{ij} - \vx_j \Vert}{\Vert \vx_i - \vx_j \Vert} u_i
  +
  \frac{\Vert \vx_{ij} - \vx_i \Vert}{\Vert \vx_i - \vx_j \Vert} u_j,
\end{align}
or the upwind interpolation
\begin{align}
  u_{ij}^\mathrm{upwind}:=
  \left\{
    \begin{array}{ll}
      u_i & \mathrm{if \ } \vc_{ij} \cdot \vn_{ij} \geq 0
      \\
      u_j & \mathrm{else}.
    \end{array}
    \right.
\end{align}
There are extensive variations of the interpolation, such as the semi-Lagrangian method and Lax--Wendroff method \cite{lax1960}.
The gradient in the direction of $\vd_{ij}:= \vx_j - \vx_i$ for the diffusion term can be computed as
\begin{align}
  \left[\nabla_{\hat{\vd}_{ij}} u\right]_{ij} \approx
  \frac{u_j - u_i}{\Vert \vd_{ij} \Vert} \hat{\vd}_{ij},
\end{align}
where
$\hat{\vd}_{ij} := \vd_{ij} / \Vert \vd_{ij} \Vert$.
Because $\vn_{ij} = \hat{\vd}_{ij}$ in the present setting, we obtain
\begin{align}
  \vn_{ij} \cdot {\left[ \nabla u \right]}_{ij}
  =
  \hat{\vd}_{ij} \cdot {\left[ \nabla_{\hat{\vd}_{ij}} u \right]}_{ij}
  \approx
  \frac{u_j - u_i}{\Vert \vd_{ij} \Vert}.
\end{align}

\subsection{Boundary Condition}
Boundary conditions are essential for solving PDEs because the behavior of the solution drastically differs depending on the conditions. We consider two types of boundary conditions: 1) the Dirichlet boundary condition defining the value of the solution on the boundary and 2) the Neumann boundary condition defining the value of the gradient of the solution on the boundary. These boundary conditions are expressed as
\begin{align}
  u(\vx) &= \hat{u}(\vx), \mathrm{\ on \ } \partial \Omega_\mathrm{Dirichlet}
  \\
  \vn \cdot [\nabla u](\vx) &= \hat{g}(\vx), \mathrm{\ on \ } \partial \Omega_\mathrm{Neumann},
\end{align}
where $\hat{\cdot}$ represents a given value, and $\partial\Omega_\mathrm{Dirichlet}$ and $\partial\Omega_\mathrm{Neumann}$ represent Dirichlet and Neumann boundaries satisfying $\partial\Omega_\mathrm{Dirichlet} \cap \partial\Omega_\mathrm{Neumann} = \emptyset$ and $\partial\Omega_\mathrm{Dirichlet} \cup \partial\Omega_\mathrm{Neumann} = \partial\Omega$, respectively. Under spatial discretization, these conditions are
\begin{align}
  u_{ij} &= \hat{u}_{ij}, \ \forall (i, j) \in \gE_\mathrm{Dirichlet}
  \\
  \vn_{ij} \cdot [\nabla u]_{ij} &= \hat{g}_{ij}, \ \forall (i, j) \in \gE_\mathrm{Neumann},
\end{align}
where $\gE_\mathrm{Dirichlet}$ and $\gE_\mathrm{Neumann}$ denote the sets of faces corresponding to the Dirichlet and Neumann boundaries, respectively. These expressions indicate that boundary conditions correspond to edge features in the GNN settings. Further, the $j$-th cell does not exist if the $i$-th cell exists and $(i, j)$ is on the boundary. Here, $j$ represents a virtual index to characterize boundary faces.

In the case of the convection term, the boundary conditions can be applied as
\begin{align}
  \begin{array}{ll}
    u_{ij} \leftarrow \hat{u}_{ij},
    &
    (i, j) \in \gE_\mathrm{Dirichlet}
    \\
    u_{ij} \leftarrow u_{ij} + \hat{g}_{ij} (\vx_{ij} - \vx_{i}) \cdot \vd_{ij},
    &
    (i, j) \in \gE_\mathrm{Neumann}
  \end{array}
\end{align}
The boundary conditions for the diffusion term are expressed as
\begin{align}
  \begin{array}{ll}
    \displaystyle
    \vn_{ij} \cdot (\nabla u)_{ij} \leftarrow \frac{\hat{u}_{ij} - u_i}{\Vert \vx_{ij} - \vx_i \Vert},
    &
    (i, j) \in \gE_\mathrm{Dirichlet}
    \\[8pt]
    \vn_{ij} \cdot (\nabla u)_{ij} \leftarrow \hat{g}_{ij},
    &
    (i, j) \in \gE_\mathrm{Neumann}
  \end{array}
\end{align}
where ``$\leftarrow$'' represents the variable overwriting during computation.

\subsection{Temporal Discretization}
We consider the temporal discretization for the general conservation form. To specify the time, we write variables explicitly as functions of time, e.g., $\vu(t)$ denoting the vector field $\vu$ at time $t$.

As is the case with spatial discretization, there are numerous methods for temporal discretization. One of the simplest schemes is the explicit Euler method expressed as
\begin{align}
  \vu(t + \Delta t) = \vu(t) - \nabla \cdot \mF(\vu) (t) \Delta t,
\end{align}
where one can compute an unknown state $\vu(t + \Delta t)$ by evaluating the right-hand side using a known state $\vu(t)$.
The other one is the implicit Euler method written as
\begin{align}
  \vu(t + \Delta t) = \vu(t) - \nabla \cdot \mF(\vu) (t + \Delta t) \Delta t.
\end{align}
The implicit method contains unknowns on both sides, requiring more expensive computations to solve the equation compared to that required by the explicit method.

\section{Proofs}
\label{app:proof}
\subsection{Proof of \cref{lem:linear}}
\begin{proof}
  We show the linearity of update functions by contradiction.
  Assume that there exists a strictly nonlinear operator $\gF_\mathrm{NL}$ that is vertex-wise, continuous, and conservative.
  Here, one can let
  $\gF_\mathrm{NL}(\vzero) = \vzero$
  without the loss of generality because of the shift-invariance of the input and output conservative fields.
  Now, consider a graph with two vertices and the vertex signal
  $\mH = (\vh_1, \vh_2)^\top$.
  Applying the vertex-wise nonlinear operator to the signal, the converted conservation constant $\tilde{\mC}$ can be expressed as
  \begin{align}
    \gF_\mathrm{NL}(\vh_1) + \gF_\mathrm{NL}(\vh_2) = \tilde{\mC}.
  \end{align}
  Converting a different input $(\vh_1 + \vh_2, \vzero)^\top$ results in
  \begin{align}
    \gF_\mathrm{NL}(\vh_1 + \vh_2) + \gF_\mathrm{NL}(0) = \tilde{\mC}.
  \end{align}
  Using $\gF_\mathrm{NL}(\vzero) = \vzero$, one can find that
  \begin{align}
    \gF_\mathrm{NL}(\vh_1) + \gF_\mathrm{NL}(\vh_2) = \gF_\mathrm{NL}(\vh_1 + \vh_2),
  \end{align}
  which results in the Cauchy's functional equation.
  The solution of the equation is known to be linear if $\gF_\mathrm{NL}$ is continuous (at least one point) \cite{kannappan2009functional}, which is a contradiction.
\end{proof}

\begin{remark}
  If a vertex-wise continuous conservative operator $\gF$ is a map to the same space,  $\gF$ is an identity because $\gF(\vh_1)$ should be conservative in a graph with one vertex.
  \label{rem:identity}
\end{remark}

\subsection{Proof of \cref{thm:locally_conservative_gnn}}
\begin{proof}
  By \cref{rem:identity}, the update function is the identity. Thus, MPNN is expressed as
  \begin{align}
    \tilde{\vh}_i = \vh_i + \sum_{j \in \gN_i} \vm_{ij}.
  \end{align}
  The condition for conservation is $\sum_{i \in \gV} \vh_i = \sum_{i \in \gV} \tilde{\vh}_i$, which results in
  \begin{align}
    \sum_{i \in \gV} \sum_{j \in \gN_i} \vm_{ij} = \vzero.
    \label{eq:condition_locally_conservative}
  \end{align}
  From the equation, $\vm_{ii} = \vzero$ for all $i \in \gV$.
  Considering the complete graph with two vertices, we see $\vm_{12} = - \vm_{21}$ from \cref{eq:condition_locally_conservative}. Now, assume that the GNN is conservative for graphs with $K$ edges. Adding the $(K+1)$-th edge $(i, j)$ requires $\vm_{ij} = - \vm_{ji}$ to retain conservation. The converse can be verified easily.
\end{proof}

\subsection{Proof of \cref{thm:encode_process_decode}}
\begin{proof}
  Following \cref{lem:linear}, encoders and decodes must be linear.
  We show that the decoder should be the left inverse of the encoder. Consider a graph with one vertex with the vertex signal $\mH = (\vh_1)^\top$.
  With that graph,
  \begin{align}
    \sum_{i \in \gV} \vh_i = \vh_1 = \mC,
  \end{align}
  where $\mC$ represents a conservation constant.
  By applying a conservative encode-process-decode architecture
  $\gF_\mathrm{decode} \circ \gF_\mathrm{L} \circ \gF_\mathrm{encode}$,
  \begin{align}
    \gF_\mathrm{decode} \circ \gF_\mathrm{L} \circ \gF_\mathrm{encode}(\vu_i)
    \nonumber
    &=
    \gF_\mathrm{decode} \circ \gF_\mathrm{L} \circ \gF_\mathrm{encode}(\mC)
    \nonumber
    \\
    &=
    \gF_\mathrm{decode} \circ \gF_\mathrm{L}(\tilde{\mC})
    \nonumber
    \\
    &=
    \gF_\mathrm{decode}(\tilde{\mC}),
    \label{eq:pseudoinverse}
  \end{align}
  where $\tilde{\mC}:= \gF_\mathrm{encode}(\mC)$ represents the conservation constant in the encoded space.
  Because of conservation, \cref{eq:pseudoinverse} is equal to $\mC$. Therefore,
  \begin{align}
    \gF_\mathrm{decode}(\tilde{\mC})
    = \gF_\mathrm{decode}(\gF_\mathrm{encode}(\mC)) = \mC.
  \end{align}
  This relationship holds for any $\mC$. Therefore, $\gF_\mathrm{decode}$ should be the left inverse of $\gF_\mathrm{encode}$.
\end{proof}

\subsection{Proof of \cref{thm:fluxgnn}}
\begin{proof}
  First, the condition for update functions (\cref{eq:locally_conservative_update}) can be satisfied by defining $\vg_i:= V_i \vh_i$.
  The message function of \cref{eq:fluxgnn} is expressed as
  \begin{align}
    \vm_{ij} = - S_{ij} \vn_{ij} \cdot \mF_\mathrm{ML} \left(\frac{1}{V_i}\vg_i, \frac{1}{V_j}\vg_{j}, \frac{1}{V_{ij}}\vg_{ij} \right),
  \end{align}
  where $V_{ij} = V_{ji}$ represents the interpolated volume at the face $(i, j)$.
  Computing $\vm_{ji}$ results in
  \begin{align}
    \vm_{ji}
    &=
    - S_{ji} \vn_{ji} \cdot \mF_\mathrm{ML} \left(\frac{1}{V_j}\vg_j, \frac{1}{V_i}\vg_i, \frac{1}{V_{ji}}\vg_{ji} \right)
    \\
    &=
    S_{ij} \vn_{ij} \cdot \mF_\mathrm{ML} \left(\frac{1}{V_j}\vg_j, \frac{1}{V_i}\vg_i, \frac{1}{V_{ij}}\vg_{ij} \right)
  \end{align}
  The condition for conservation (\cref{eq:locally_conservative_message}) requires $\vm_{ij} = - \vm_{ji}$ for all $(i, j) \in \gE$, and therefore,
  \begin{align}
    \mF_\mathrm{ML} \left(\vh_i, \vh_{j}, \vh_{ij} \right)
    =
    \mF_\mathrm{ML} \left(\vh_j, \vh_i, \vh_{ij} \right).
  \end{align}
  The converse can be shown similarly.
\end{proof}
\begin{remark}
  With FluxGNN, $\mG:= (\vg_1, \vg_2, \dots, \vg_{\VV})$ is conservative rather than $\mH$, which means that
  \begin{align}
    \sum_{j \in \gV} \vg_i = \sum_{j \in \gV} V_i \vh_i \approx \int_\Omega \vh \ dV
  \end{align}
  is conservative as with FVM.
\end{remark}

\section{FluxGNN Details}
\label{app:fluxgnn}

\subsection{Validation of FVM}
\label{app:validation}
Our model relies considerably on FVM implementation, and therefore, validating the implemented FVM is critical for successfully modeling FluxGNN\@. Here, we introduce the results of the validation.

\subsubsection*{Convection--Diffusion Equation}
We performed the study under the same condition as the machine learning task, i.e., periodic boundary condition with a sinusoidal initial condition. \cref{tab:validation_cd} and \cref{tab:validation_cd} present the results of the convergence study in the convection--diffusion equation. We confirmed that the convergence is in the second order for the spatial resolution, as expected from the formulation of FVM.

\begin{figure}[bt]
  \centering
  \centerline{\includegraphics[trim={0cm 0cm 0cm 0cm},width=0.5\linewidth]{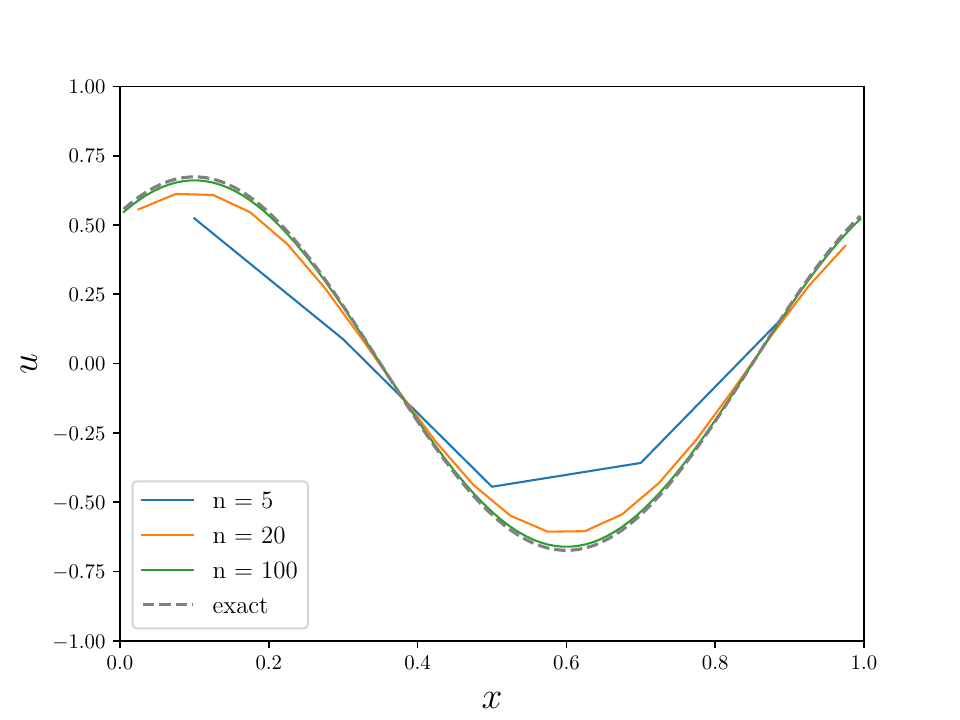}}
    \caption{Results of the convergence study of FVM for the convection--diffusion equation at $t = 1.0$.}
  \label{fig:validation_cd}
\end{figure}

\begin{table}[bt]
  \caption{Results of the convergence study of FVM for the convection--diffusion equation.}
  \centering
  \label{tab:validation_cd}
  \begin{small}
    \begin{tabular}{ccc}
      \toprule
      $n_\mathrm{cell}$
      &
      $\Delta x$
      &
      RMSE
      \\
      \midrule
      5 & 0.2 & 0.128
      \\
      10 & 0.1 & 0.077
      \\
      20 & 0.05 & 0.043
      \\
      50 & 0.02 & 0.018
      \\
      100 & 0.01 & 0.009
      \\
      \bottomrule
    \end{tabular}
  \end{small}
\end{table}

\subsubsection*{Navier--Stokes Equations}
\begin{figure}[bt]
  \centering
  \centerline{\includegraphics[trim={0cm 0cm 0cm 0cm},width=0.5\linewidth]{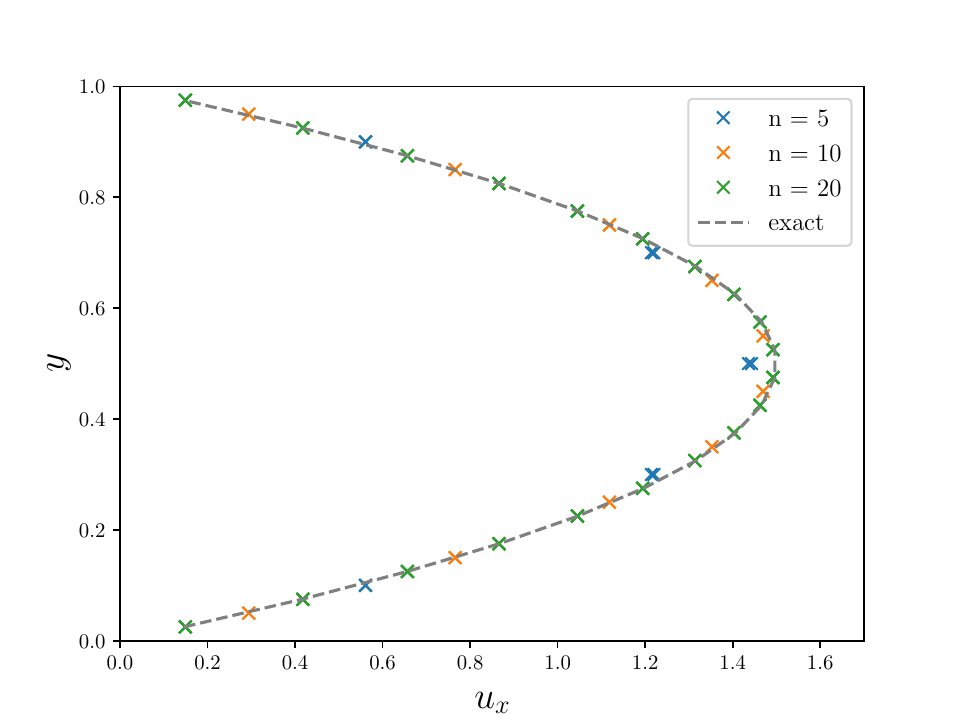}}
  \caption{Results of the convergence study of FVM for the Navier--Stokes equations with the 2D Hagen–-Poiseuille flow at the steady state.}
  \label{fig:validation_ns}
\end{figure}

We performed a convergence study of the Navier--Stokes equation for the 2D Hagen--Poiseuille flow. We generated a rectangular shape with the length in the $x$ and $y$ directions of two and one, respectively. The inlet is on the side with the minimum $x$ coordinate. The magnitude of inflow is one, and the Reynolds number is one. Under this setting, the $x$ component of the velocity as a function of the $y$ coordinate, $u_x(y)$, is given as
\begin{align}
  u_x(y) = 6 y (1 - y).
\end{align}
\cref{fig:validation_ns} and \cref{tab:validation_ns} present the results, which re-confirm the second-order convergence.

\begin{table}[bt]
  \caption{Results of the convergence study of FVM for the Navier--Stokes equations with the 2D Hagen–-Poiseuille flow.}
  \centering
  \label{tab:validation_ns}
  \begin{small}
    \begin{tabular}{ccc}
      \toprule
      $n_\mathrm{cell}$ in the $y$ direction
      &
      $\Delta x$
      &
      RMSE
      \\
      \midrule
      5 & 0.2 & 0.041
      \\
      10 & 0.1 & 0.010
      \\
      20 & 0.05 & 0.003
      \\
      \bottomrule
    \end{tabular}
  \end{small}
\end{table}

\subsubsection*{Navier--Stokes Equations with Mixture}
\begin{figure}[bt]
  \centering
  \centerline{\includegraphics[trim={0cm 2cm 0cm 0cm},width=0.9\linewidth]{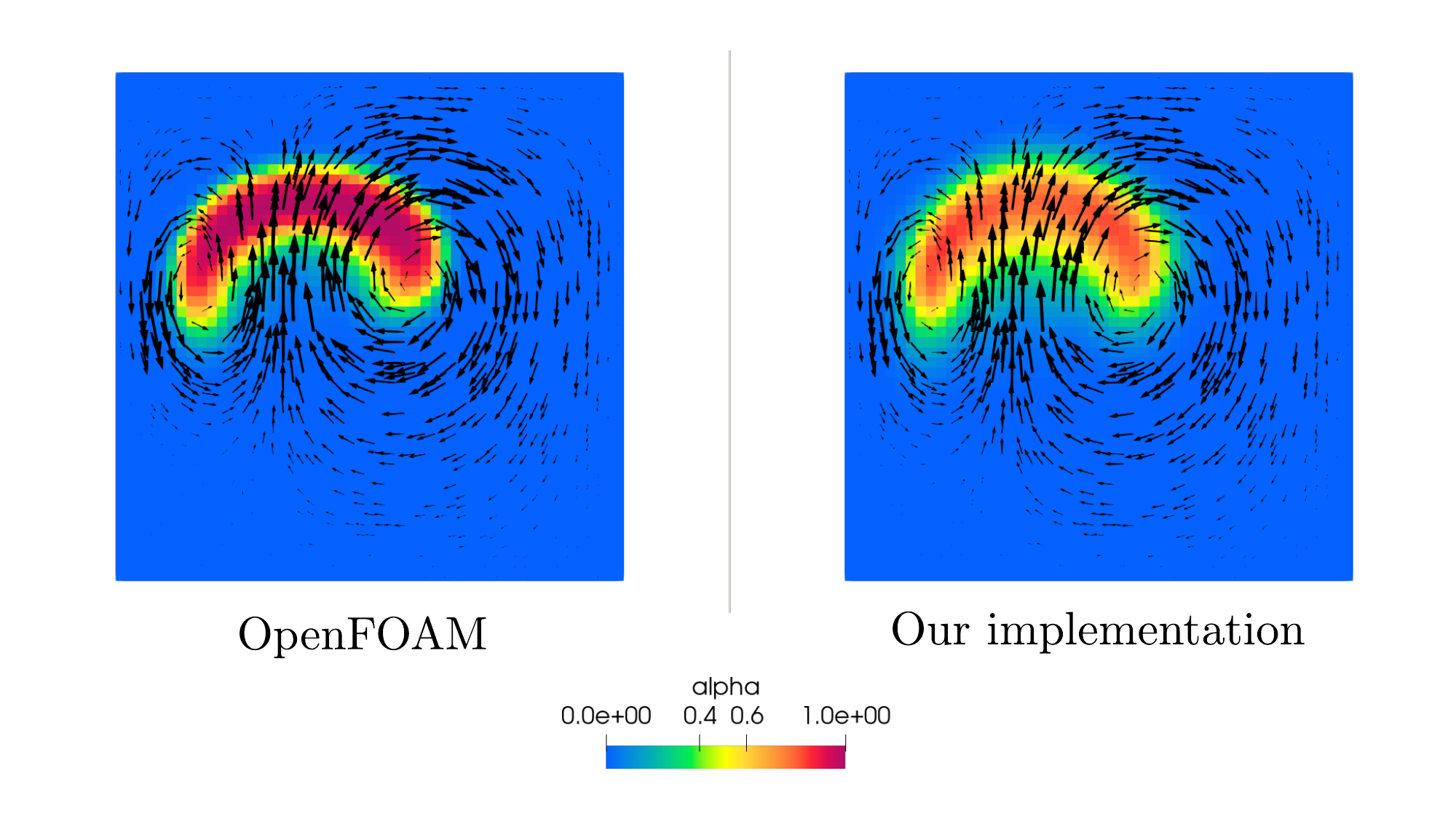}}
    \caption{Results of the validation of our FVM for the Navier--Stokes equation with mixture. Arrows indicate the velocity fields.}
  \label{fig:validation_mixture}
\end{figure}

We conducted a test of our FVM using the Navier--Stokes equations with mixture. The experimental setup closely mirrors that of the machine learning task, with the exception of the diffusion coefficient for $\alpha$, denoted as $D_\alpha$, which has been set to $10^{-4}$ for stability reasons. Constructing an exact solution for this class of problems is challenging, and therefore, we opted to compare our numerical solution with that of OpenFOAM, a well-established classical solver that employs FVM\@. \cref{fig:validation_mixture} presents the qualitative comparison between OpenFOAM and our FVM\@ implementation. While our implementation tends to exhibit slightly higher diffusivity at this resolution, both implementations display similar behavior.

\subsection{Flux Function}
\label{app:flux_function}
Flux functions $\mF_\mathrm{ML}$ form the core of FluxGNNs. Here, we present definitions of the flux functions used for the following experiments. The flux function for convection, $\mF_\mathrm{conv}$, is defined as
\begin{align}
  \mF_\mathrm{conv}(\tilde{u}_i, \tilde{u}_j, \tilde{u}_{ij}, \tilde{\vc}_{ij}):= \vf_\mathrm{sim} \left(
  \tilde{\vc}_{ij} F_\mathrm{int}(\tilde{u}_i, \tilde{u}_j, \tilde{u}_{ij})
  \right),
\end{align}
where $\vf_\mathrm{sim}$, $\tilde{\cdot}$, and $F_\mathrm{int}$ respectively represent a similarity-equivariant MLP defined in \cref{eq:similarity_equivariant_mlp}, feature encoded using a linear map, and flux function for interpolation.
$F_\mathrm{int}$ is expressed as
\begin{align}
  F_\mathrm{int}(\tilde{u}_i, \tilde{u}_j, \tilde{u}_{ij})
 :=
  \frac{1}{4}\left[
    f_\mathrm{sim}^\mathrm{vertex}(\tilde{u}_i) + f_\mathrm{sim}^\mathrm{vertex}(\tilde{u}_j)
    + f_\mathrm{sim}^\mathrm{linear}(\tilde{u}_{ij}^\mathrm{linear})
    + f_\mathrm{upwind}^\mathrm{vertex}(\tilde{u}_{ij}^\mathrm{upwind})
    \right],
\end{align}
where $\tilde{u}_{ij} = (\tilde{u}_{ij}^\mathrm{linear}, \tilde{u}_{ij}^\mathrm{upwind})$ is computed using FVM schemes explained in \cref{app:fvm}. The similarity-equivariant MLP for vertex signals is shared to achieve permutation-equivariance for vertex signals. However, this does not apply to the MLPs for edge signals.

The gradient for the diffusion term is modeled as
\begin{align}
  \mF_\mathrm{grad}(\tilde{u}_i, \tilde{u}_j, \tilde{u}_{ij})
 :=
  \vf_\mathrm{sim} \left(
    \frac{\tilde{u}_j - \tilde{u}_i}{\Vert \vd_{ij} \Vert}
    \frac{\vd_{ij}}{\Vert \vd_{ij} \Vert}
    \right).
\end{align}

The boundary conditions are applied using the same procedure as \cref{app:fvm}; however, in the encoded space.

\section{Experiment Details: Convection--Diffusion Equation}
\label{app:cd}

\subsection{Dataset}
The equation of interest is
\begin{align}
  \pdiff{}{t} u = - \nabla \cdot \left(\vc u - D \nabla u \right),
  \ \ (t, x) \in (0, T_\mathrm{max}) \times [0, 1),
\end{align}
with periodic boundary conditions, where $D = 10^{-4}$ and $\Vert \vc \Vert \in [0.0, 0.2]$ is obtained from the uniform distribution. We define the initial condition as
\begin{align}
  u(t = 0, x) = u_\mathrm{amp} \cos (2 \pi (x + x_0)), \ \ x \in [0.0, 1.0)
\end{align}
with $u_\mathrm{amp} \in [0.5, 1.0]$ and $x_0 \in [0.0, 1.0]$ obtained from the uniform distribution. A precise solution to the specified initial condition, when subjected to periodic boundary conditions is
\begin{align}
  u(t, x) = u_\mathrm{amp} \exp(- (2 \pi)^2 D t) \cos(2 \pi (x - ct + x_0)).
\end{align}
We generated the dataset by randomly varying the initial condition and convecting velocity.

\subsection{FluxGNN Modeling}
The FluxGNN model used for the convection--diffusion equation is expressed as explained below.

\subsubsection*{Encoder}
All encoders are linear following \cref{thm:encode_process_decode}. Using vertex-wide linear functions $\{\epsilon_\cdot\}$, each input feature is encoded as
\begin{align}
  \begin{array}{rcl}
    u:
    &
    \tilde{u}_i = \epsilon_u(u_i),
    &
    i \in \gV
    \\
    \mathrm{Dirichlet \ for \ } u:
    &
    \tilde{\hat{u}}_{ij} = \epsilon_u(\hat{u}_{ij}),
    &
    (i, j) \in \gE_\mathrm{Dirichlet}
    \\
    \mathrm{Neumann \ for \ } u:
    &
    \tilde{\hat{g}}_{ij} = \epsilon_u(\hat{g}_{ij}),
    &
    (i, j) \in \gE_\mathrm{Neumann}
    \\
    \vc:
    &
    \tilde{\vc}_{ij} = \epsilon_\vc(\vc_{ij}),
    &
    (i, j) \in \gE
    \\
    D:
    &
    \tilde{D}_{ij} = \epsilon_D(D_{ij}),
    &
    (i, j) \in \gE
    \\
  \end{array}
\end{align}

We used encoders with embedded dimensions of 64.

\subsubsection*{Processor}
The processing part is constructed using FVM formulation and FluxGNN as
\begin{align}
  \tilde{u}(t + \Delta t)
  &=
  \tilde{u}(t)
  +
  \sum_{j \in \gN_i} S_{ij} \vn_{ij} \cdot \left[
    \mF_\mathrm{conv}(\tilde{u}_i, \tilde{u}_j, \tilde{u}_{ij}, \tilde{\vc}_{ij})
    - \tilde{D}_{ij} \mF_\mathrm{grad}(\tilde{u}_i, \tilde{u}_j, \tilde{u}_{ij})
    \right] \Delta t,
\end{align}
where we apply the explicit Euler scheme for time evolution.
For MLPs used in similarity-equivariant MLPs, we used the structure
\begin{align}
  \mathrm{MLP}(\vh):= \tanh \circ \mathrm{Linear}_{64 \leftarrow 64}\supp{2} \circ \tanh \circ \mathrm{Linear}_{64 \leftarrow 64}\supp{1}(\vh),
\end{align}
where $\mathrm{Linear}_{64 \leftarrow 64}\supp{\cdot}$ denotes a linear layer mapping from 64 features to 64 features.

\subsubsection*{Decoder}
We obtain a time series of encoded $u$ using the autoregressive computation in the processing step. Applying the decoder $\delta_u$ that satisfies $\delta_u \circ \epsilon_u = \mathrm{id}$, we obtain
\begin{align}
  u_i^\mathrm{pred}(t):= \delta_u(\tilde{u}_i(t)), \ \ \forall i \in \gV, \forall t \in \{\Delta t, 2 \Delta t, \dots, T_\mathrm{max}\}.
\end{align}

\section{Experiment Details: Navier--Stokes Equations with Mixture}
\label{app:mixture}

\subsection{Dataset}

\begin{figure}[bt]
  \centering
  \centerline{\includegraphics[trim={0cm 0cm 20cm 0cm},width=0.3\linewidth]{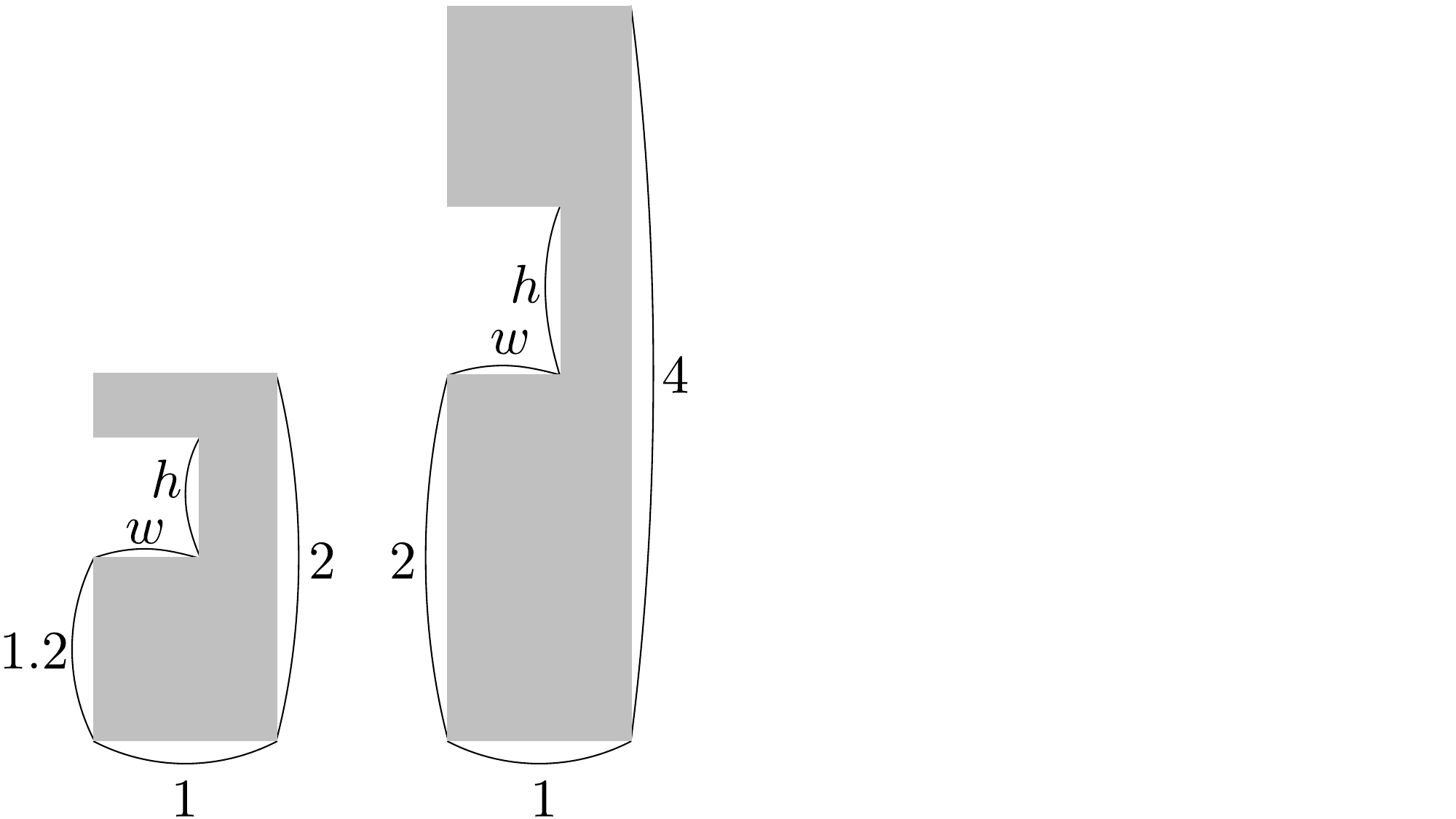}}
    \caption{Template shapes for training, validation, and test dataset (left) and taller dataset (right). Both use two parameters, $h$ and $w$.}
  \label{fig:mixture_shape}
\end{figure}

We generated shapes with varying dimensions, as shown in \cref{fig:mixture_shape}. Specifically, we varied $h$ and $w$ across these settings: 0.2, 0.4, and 0.6, resulting in a total of nine distinct shapes. To avoid data leakage, we partitioned these shapes into three sets: five for training, two for validation, and two for testing. Subsequently, we created four initial conditions for each shape using random numerical analysis. These initial conditions were employed in OpenFOAM simulations, utilizing fine meshes with a spatial resolution of $\Delta x = 1 / 320$. Additionally, we configured the simulation with the following parameters: kinematic viscosity $\nu = 10^{-3}$, diffusion coefficient $D_\alpha = 10^{-6}$, density of solvent $\rho_\mathrm{high} = 1000$, density of solute $\rho_\mathrm{low} = 990$, and gravity acting in the $y$ direction with a magnitude of 9.81. We employed the isolated boundary condition with walls, specifically:
\begin{align}
  \begin{array}{lll}
    \vu(\vx) = \vzero, & \mathrm{on} & \partial\Omega
    \\
    p(\vx) = 0, & \mathrm{on} & \partial\Omega_{\min\! y}
    \\
    \vn(\vx) \cdot \nabla p(\vx) = 0, & \mathrm{on} & \partial\Omega \setminus \partial\Omega_{\min\! y}
    \\
    \vn(\vx) \cdot \nabla \alpha(\vx) = 0, & \mathrm{on} & \partial\Omega
  \end{array}
\end{align}
where $\partial\Omega_{\min\! y}$ represents the boundary at the bottom.

After OpenFOAM simulations, we interpolated the results to coarsened meshes with $\Delta x = 1 / 20$. This was done deliberately to create a more demanding machine learning problem because coarsened meshes are known to introduce significant numerical diffusion, resulting in difficulty in maintaining the density field sharp. Our primary objective with the machine learning task is intended to examine the ability of each method to effectively maintain the sharpness of the density field (equivalently, $\alpha$), even when faced with pronounced numerical diffusion.

We divided the time series on coarse meshes into 100 trajectories with shifting time windows whose width and step size are 1.6 to augment the training dataset. Finally, we obtain 200, 8, and 8 trajectories for the training, validation, and test datasets.

We generated a rotation dataset by applying random rotation and a scaling dataset by applying a random scaling of space, time, and mass scales to evaluate the generalizability of each method. Further, we generated the taller dataset, which has taller shapes than the test dataset, as illustrated in \cref{fig:mixture_shape}. We chose dimension parameters as $(h, w) = (0.5, 0.2), (0.5, 0.4), \text{and } (0.5, 0.6)$. We generated one trajectory for each shape in the taller dataset using the same procedure to generate the test dataset.

\subsection{FluxGNN modeling}
The fundamental structure of FluxGNN remain consistent with the approach applied for the convection--diffusion equation, with the addition of established machine learning techniques, including temporal bundling and the neural nonlinear solver. The hyperparameter employed for the study is detailed in \cref{tab:mixture_fluxgnn_hyperparameters}.

To obtain the pressure field, we applied the fractional step method as done in \cite{horie2022physics}. The equations to solve are turned into
\begin{align}
  \pdiff{}{t}\alpha
  &=
  - \nabla \cdot \left[
    \vu \alpha - D_\alpha \nabla \alpha
    \right]
  \\
  \widetilde{\rho \vu}:&= \rho \vu - \nabla \cdot \left[
    \rho \vu \otimes \vu
    \frac{\nu}{\rho} \left[ \nabla \otimes \vu - (\nabla \otimes \vu)^\top \right]
    \right] \Delta t
  \\
  \nabla \cdot \nabla (p_{\rho gy}^+)
  &=
  - \nabla \cdot (\vg \cdot \vh \nabla \rho) + \frac{1}{\Delta t}\left[
    \nabla \cdot \widetilde{\rho \vu} + \pdiff{}{t}\rho
    \right]
  \\
  \vu^+
  &=
  \frac{1}{\rho} \left[
    \widetilde{\rho \vu} - \Delta t (\nabla p^+_{\rho gh} + \vg \cdot \vh \nabla \rho)
    \right],
\end{align}
where $\cdot^+$ denotes variables at the next time step.
We utilized a matrix-free conjugate gradient (CG) method \cite{prabhune2020fast}, which corresponds to an iterative application of locally conservative diffusion operations. In the CG method, we refrain from using trainable functions to solve the Poisson equation for pressure. This choice is made due to the potentially large number of iterations required (e.g., 20 to 100 iterations), as the presence of trainable parameters in a deep loop often leads to instability in the backward process.

\begin{table}[bt]
  \caption{Hyperparameter range used for FluxGNN.}
  \centering
  \label{tab:mixture_fluxgnn_hyperparameters}
  \begin{small}
    \begin{tabular}{ll}
      \toprule
      Name
      &
      Range
      \\
      \midrule
      $n_\mathrm{bundle}$: \# of steps bundled
      &
      2, 4
      \\
      $n_\mathrm{feature}$: \# encoded features
      &
      8, 16, 32, 64
      \\
      $n_\mathrm{rep}$: \# neural nonlinear steps
      &
      2, 4, 8
      \\
      \bottomrule
    \end{tabular}
  \end{small}
\end{table}

\subsection{Baselines}
The visibility of a wide range of hops in the model is attributed to FluxGNN's utilization of numerous iterations within the forward loop. Consequently, our hyperparameter investigations encompassed baseline models with varying levels of visible hops, in addition to exploring different configurations of hidden features. The comprehensive details of these study parameters can be found in \cref{tab:mixture_mppde_hyperparameters,tab:mixture_penn_hyperparameters,tab:mixture_fvm_hyperparameters}.

\begin{table}[tbh]
  \caption{Hyperparameter range used for MP-PDE.}
  \centering
  \label{tab:mixture_mppde_hyperparameters}
  \begin{small}
    \begin{tabular}{ll}
      \toprule
      Name
      &
      Range
      \\
      \midrule
      $n_\mathrm{bundle}$: \# of steps bundled
      &
      2, 4, 8
      \\
      $n_\mathrm{feature}$: \# encoded features
      &
      32, 64, 128
      \\
      $n_\mathrm{neighbor}$: \# neighbors considered in one forward computation of a GNN layer
      &
      4, 8, 16
      \\
      \bottomrule
    \end{tabular}
  \end{small}
\end{table}

\begin{table}[tbh]
  \caption{Hyperparameter range used for PENN.}
  \centering
  \label{tab:mixture_penn_hyperparameters}
  \begin{small}
    \begin{tabular}{ll}
      \toprule
      Name
      &
      Range
      \\
      \midrule
      $n_\mathrm{feature}$: \# encoded features
      &
      4, 8, 16, 32, 64
      \\
      $n_\mathrm{rep}$: \# neural nonlinear steps
      &
      4, 8, 16
      \\
      \bottomrule
    \end{tabular}
  \end{small}
\end{table}

\begin{table}[tbh]
  \caption{Hyperparameter range used for FVM.}
  \centering
  \label{tab:mixture_fvm_hyperparameters}
  \begin{small}
    \begin{tabular}{ll}
      \toprule
      Name
      &
      Range
      \\
      \midrule
      $n_{\mathrm{rep} \vu}$: \# loops for solver of $\vu$
      &
      4, 8, 16, 32, 64, 128
      \\
      $n_{\mathrm{rep} \alpha}$: \# loops for solver of $\alpha$
      &
      4, 8, 16, 32, 64, 128
      \\
      \bottomrule
    \end{tabular}
  \end{small}
\end{table}

\subsection{Results}
\begin{figure}[tb]
  \centering
  \centerline{\includegraphics[trim={0cm 5cm 0cm 0cm},width=0.95\linewidth]{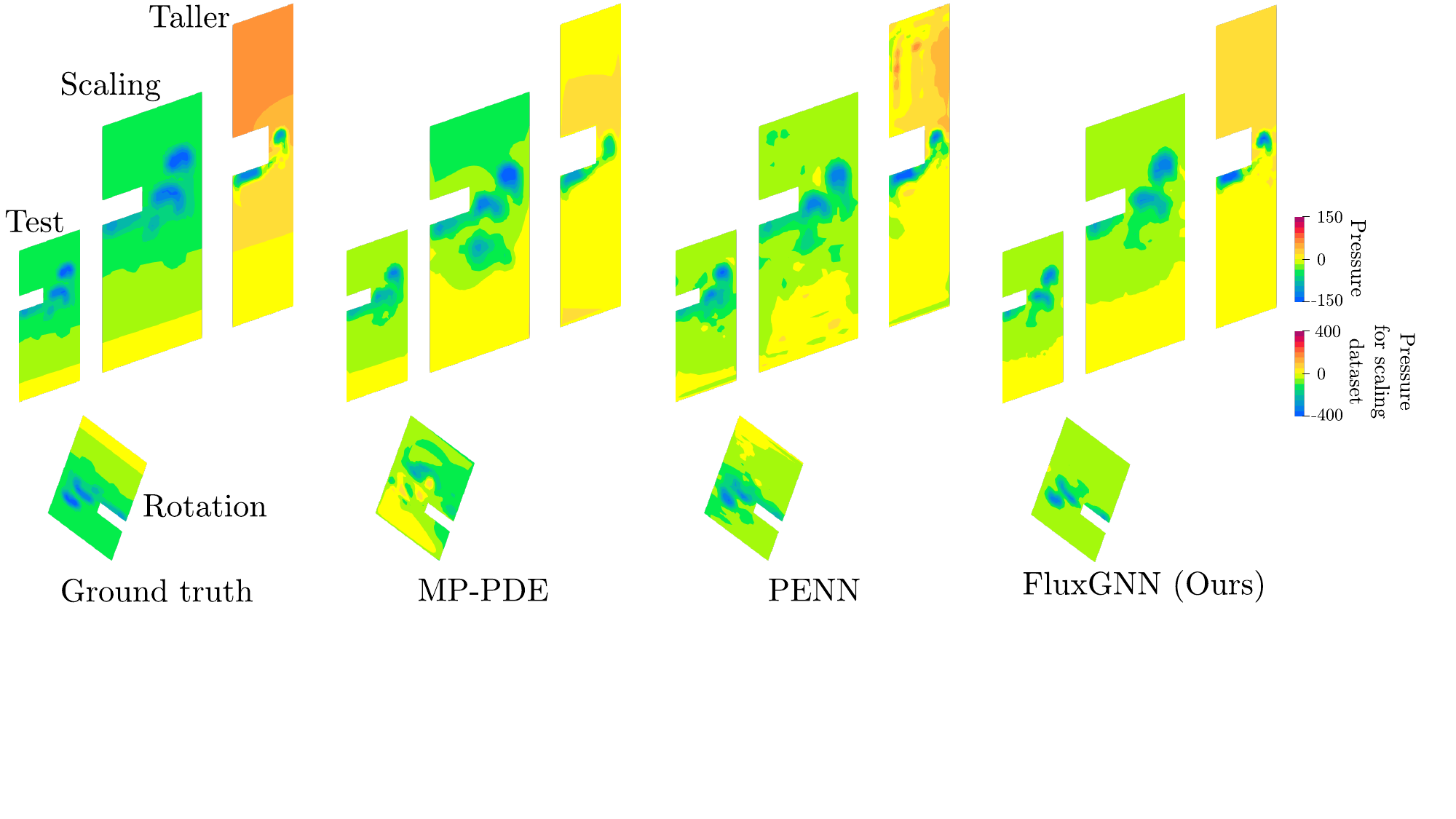}}
  \caption{Visual comparison of pressure field between ground truth, MP-PDE, PENN, and FluxGNN.}
  \label{fig:mixture_p}
\end{figure}

\begin{figure}[tb]
  \centering
  \centerline{\includegraphics[trim={0cm 5cm 0cm 0cm},width=0.95\linewidth]{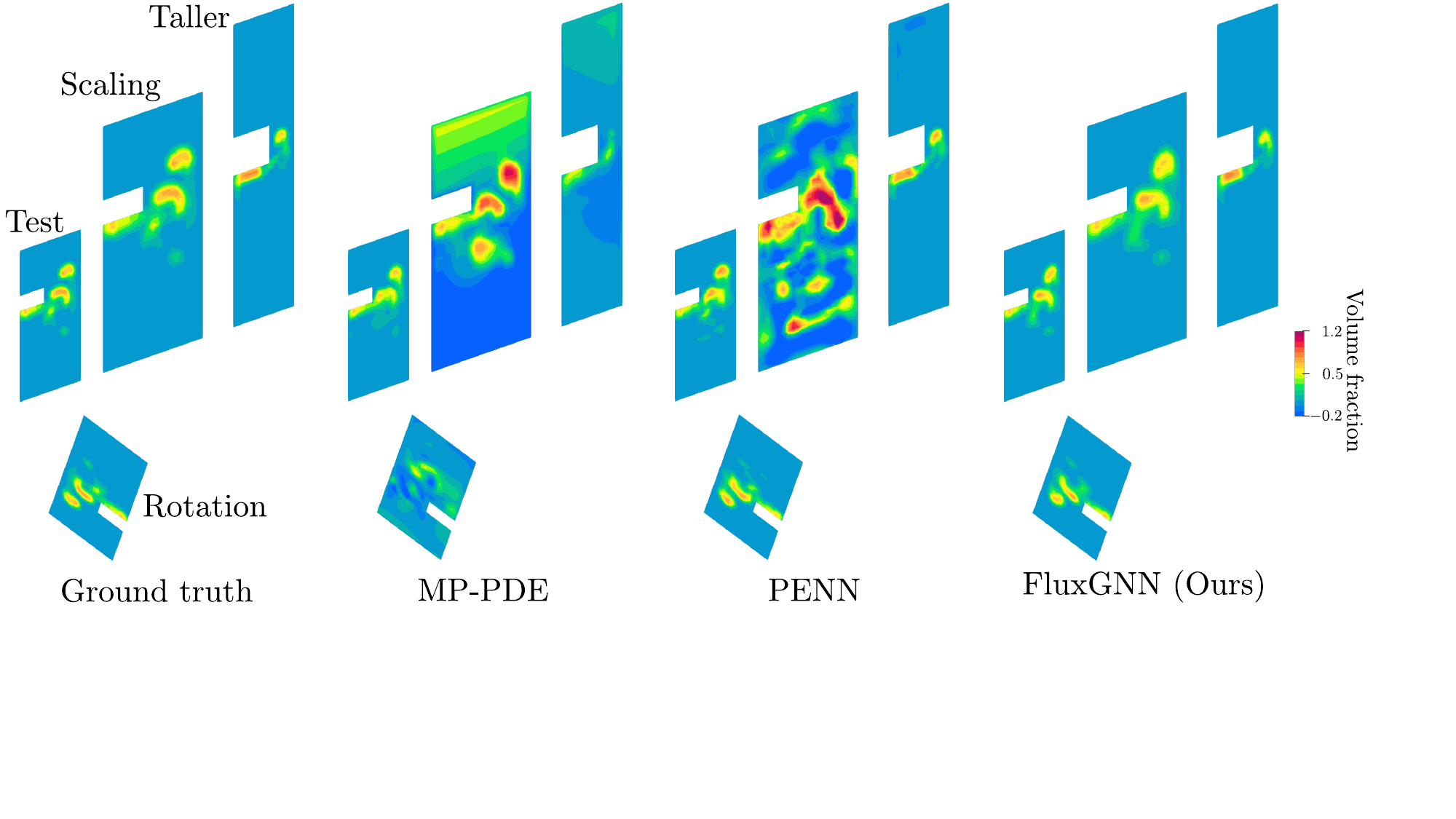}}
  \caption{Visual comparison of volume fraction field between ground truth, MP-PDE, PENN, and FluxGNN.}
  \label{fig:mixture_alpha}
\end{figure}

\cref{fig:mixture_p,fig:mixture_alpha} present additional visualizations of the results. Notably, FluxGNN exhibits robust performance, even when confronted with unseen shapes from the taller dataset. Interestingly, PENN also demonstrates proficient performance in predicting $\alpha$ for these unfamiliar shapes, likely owing to the $\En$-equivariance embedded within the model.

\begin{figure}[bt]
  \centering
  \centerline{\includegraphics[trim={0cm 0cm 0cm 0cm},width=0.9\linewidth]{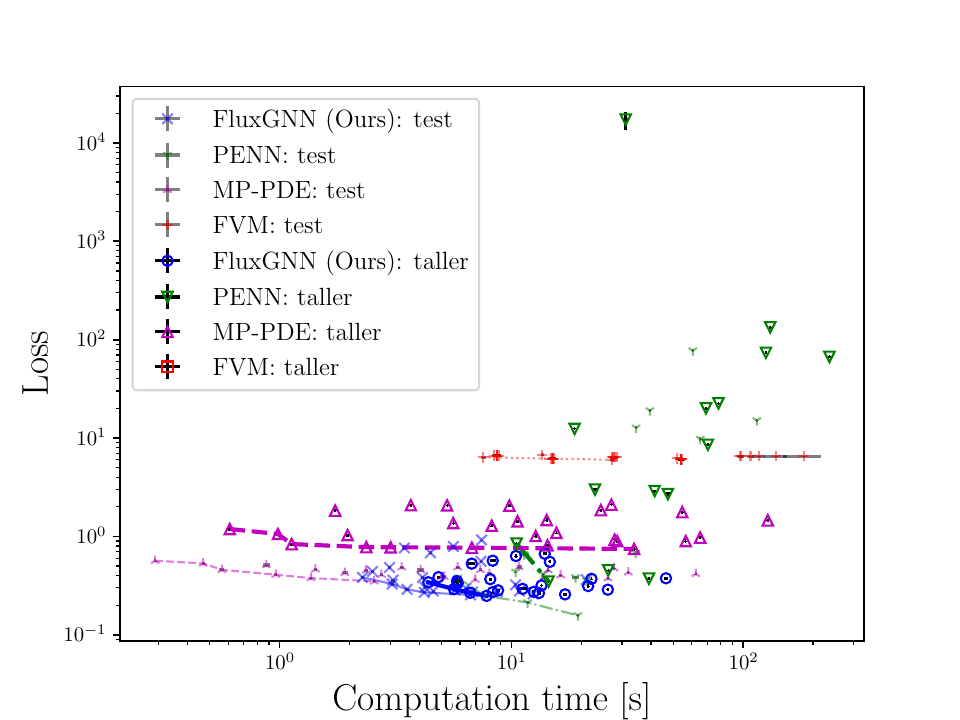}}
    \caption{Speed--accuracy tradeoff of machine learning models (MP-PDE, PENN, and FluxGNN) and FVM, with error bars corresponding to the standard error of the mean. Light and dark colors correspond to the evaluation of the test and taller datasets, respectively. Lines represent Pareto fronts. FVM did not converge on the taller dataset with all settings tested, and therefore, there is no plot for FVM on the dataset. All computations are done on the same CPU (Intel Xeon CPU E5-2695 v2 @ 2.40 GHz) with one core.}
  \label{fig:mixture_tradeoff_detail}
\end{figure}

\cref{fig:mixture_tradeoff_detail} provides a concise overview of the tradeoff between speed and accuracy, as ascertained through hyperparameter studies. All data used for the plot are shown in \cref{tab:fvm_hyper,tab:mppde_hyper,tab:penn_hyper,tab:fluxgnn_hyper}. In the tested domain, our model consistently achieves higher accuracy compared to MP-PDE, primarily because it incorporates the underlying laws of physics. Meanwhile, MP-PDE showcases commendable computational efficiency due to its less specific physics implementation, resulting in a simpler overall model structure. PENN, on the other hand, delivers superior accuracy compared to MP-PDE models but incurs a relatively longer runtime, as it lacks dedicated mechanisms for time-series computation. Our FluxGNN models have high generalizability, as demonstrated in \cref{sec:experiments}. Furthermore, they tend to achieve higher computational efficiency than PENN, owing to the incorporation of temporal bundling, a method designed to streamline time-series computation.

\begin{table}[bt]
  \caption{Detailed results of the hyperparameter study for FVM\@. All computations are done on the same CPU (Intel Xeon CPU E5-2695 v2 @ 2.40 GHz) with one core.}
  \centering
  \label{tab:fvm_hyper}
  \begin{small}
    \begin{tabular}{rrrrrr}
      \toprule
      $n_{\mathrm{rep} \vu}$
      &
      $n_{\mathrm{rep} \alpha}$
      &
      \makecell{%
        Computation time
        \\
        on the test dataset [s]
      }
      &
      \makecell{%
        Loss
        \\
        on the test dataset
        \\
        ($\times 10^{-1}$)
      }
      &
      \makecell{%
        Computation time
        \\
        on the taller dataset [s]
      }
      &
      \makecell{%
        Loss
        \\
        on the taller dataset
        \\
        ($\times 10^{-1}$)
      }
      \\
      \midrule
4 & 4 &
$5.48 \pm 1.33$ &
$\mathrm{NaN} \pm \mathrm{NaN}$ &
$39.83 \pm 7.38$ &
$\mathrm{NaN} \pm \mathrm{NaN}$
\\
8 & 4 &
$5.81 \pm 0.97$ &
$\mathrm{NaN} \pm \mathrm{NaN}$ &
$43.83 \pm 6.65$ &
$\mathrm{NaN} \pm \mathrm{NaN}$
\\
16 & 4 &
$6.61 \pm 0.97$ &
$\mathrm{NaN} \pm \mathrm{NaN}$ &
$42.23 \pm 7.37$ &
$\mathrm{NaN} \pm \mathrm{NaN}$
\\
32 & 4 &
$7.71 \pm 1.10$ &
$\mathrm{NaN} \pm \mathrm{NaN}$ &
$46.14 \pm 7.50$ &
$\mathrm{NaN} \pm \mathrm{NaN}$
\\
64 & 4 &
$8.99 \pm 1.32$ &
$\mathrm{NaN} \pm \mathrm{NaN}$ &
$50.53 \pm 7.37$ &
$\mathrm{NaN} \pm \mathrm{NaN}$
\\
128 & 4 &
$11.06 \pm 1.78$ &
$\mathrm{NaN} \pm \mathrm{NaN}$ &
$60.31 \pm 7.36$ &
$\mathrm{NaN} \pm \mathrm{NaN}$
\\
4 & 8 &
$7.57 \pm 0.17$ &
$63.72 \pm 0.33$ &
$108.40 \pm 13.99$ &
$\mathrm{NaN} \pm \mathrm{NaN}$
\\
8 & 8 &
$8.41 \pm 0.16$ &
$66.26 \pm 0.34$ &
$98.27 \pm 15.83$ &
$\mathrm{NaN} \pm \mathrm{NaN}$
\\
16 & 8 &
$8.69 \pm 0.15$ &
$66.26 \pm 0.34$ &
$100.47 \pm 9.41$ &
$\mathrm{NaN} \pm \mathrm{NaN}$
\\
32 & 8 &
$8.66 \pm 0.18$ &
$66.26 \pm 0.34$ &
$106.87 \pm 9.77$ &
$\mathrm{NaN} \pm \mathrm{NaN}$
\\
64 & 8 &
$8.80 \pm 0.27$ &
$66.26 \pm 0.34$ &
$110.52 \pm 15.14$ &
$\mathrm{NaN} \pm \mathrm{NaN}$
\\
128 & 8 &
$8.69 \pm 0.34$ &
$66.26 \pm 0.34$ &
$125.54 \pm 15.92$ &
$\mathrm{NaN} \pm \mathrm{NaN}$
\\
4 & 16 &
$13.58 \pm 0.21$ &
$67.40 \pm 0.27$ &
$179.02 \pm 42.98$ &
$\mathrm{NaN} \pm \mathrm{NaN}$
\\
8 & 16 &
$15.02 \pm 0.34$ &
$61.77 \pm 0.26$ &
$186.34 \pm 44.78$ &
$\mathrm{NaN} \pm \mathrm{NaN}$
\\
16 & 16 &
$14.77 \pm 0.35$ &
$61.77 \pm 0.26$ &
$188.38 \pm 44.49$ &
$\mathrm{NaN} \pm \mathrm{NaN}$
\\
32 & 16 &
$15.20 \pm 0.36$ &
$61.77 \pm 0.26$ &
$196.69 \pm 45.23$ &
$\mathrm{NaN} \pm \mathrm{NaN}$
\\
64 & 16 &
$15.16 \pm 0.36$ &
$61.77 \pm 0.26$ &
$213.58 \pm 47.77$ &
$\mathrm{NaN} \pm \mathrm{NaN}$
\\
128 & 16 &
$15.07 \pm 0.35$ &
$61.77 \pm 0.26$ &
$242.97 \pm 51.98$ &
$\mathrm{NaN} \pm \mathrm{NaN}$
\\
4 & 32 &
$27.13 \pm 0.47$ &
$60.10 \pm 0.25$ &
$338.46 \pm 128.31$ &
$\mathrm{NaN} \pm \mathrm{NaN}$
\\
8 & 32 &
$27.63 \pm 0.46$ &
$63.99 \pm 0.26$ &
$342.37 \pm 127.92$ &
$\mathrm{NaN} \pm \mathrm{NaN}$
\\
16 & 32 &
$27.51 \pm 0.44$ &
$63.99 \pm 0.26$ &
$350.57 \pm 130.42$ &
$\mathrm{NaN} \pm \mathrm{NaN}$
\\
32 & 32 &
$27.24 \pm 0.48$ &
$63.99 \pm 0.26$ &
$365.23 \pm 135.17$ &
$\mathrm{NaN} \pm \mathrm{NaN}$
\\
64 & 32 &
$28.47 \pm 0.47$ &
$63.99 \pm 0.26$ &
$392.99 \pm 145.17$ &
$\mathrm{NaN} \pm \mathrm{NaN}$
\\
128 & 32 &
$27.94 \pm 0.46$ &
$63.99 \pm 0.26$ &
$441.77 \pm 164.01$ &
$\mathrm{NaN} \pm \mathrm{NaN}$
\\
4 & 64 &
$51.99 \pm 1.29$ &
$62.07 \pm 0.25$ &
$810.78 \pm 202.26$ &
$\mathrm{NaN} \pm \mathrm{NaN}$
\\
8 & 64 &
$54.14 \pm 1.49$ &
$60.69 \pm 0.25$ &
$771.13 \pm 183.60$ &
$\mathrm{NaN} \pm \mathrm{NaN}$
\\
16 & 64 &
$53.82 \pm 1.46$ &
$60.69 \pm 0.25$ &
$834.60 \pm 205.83$ &
$\mathrm{NaN} \pm \mathrm{NaN}$
\\
32 & 64 &
$53.93 \pm 1.51$ &
$60.69 \pm 0.25$ &
$862.31 \pm 210.42$ &
$\mathrm{NaN} \pm \mathrm{NaN}$
\\
64 & 64 &
$54.43 \pm 1.50$ &
$60.69 \pm 0.25$ &
$923.02 \pm 222.67$ &
$\mathrm{NaN} \pm \mathrm{NaN}$
\\
128 & 64 &
$53.76 \pm 1.54$ &
$60.69 \pm 0.25$ &
$1088.87 \pm 267.00$ &
$\mathrm{NaN} \pm \mathrm{NaN}$
\\
4 & 128 &
$96.50 \pm 2.89$ &
$65.58 \pm 0.26$ &
$1830.75 \pm 371.98$ &
$\mathrm{NaN} \pm \mathrm{NaN}$
\\
8 & 128 &
$98.27 \pm 2.73$ &
$65.58 \pm 0.26$ &
$1721.97 \pm 444.38$ &
$\mathrm{NaN} \pm \mathrm{NaN}$
\\
16 & 128 &
$107.59 \pm 3.66$ &
$65.58 \pm 0.26$ &
$1749.10 \pm 446.70$ &
$\mathrm{NaN} \pm \mathrm{NaN}$
\\
32 & 128 &
$117.08 \pm 7.49$ &
$65.58 \pm 0.26$ &
$1825.21 \pm 468.47$ &
$\mathrm{NaN} \pm \mathrm{NaN}$
\\
64 & 128 &
$138.57 \pm 15.99$ &
$65.58 \pm 0.26$ &
$1915.15 \pm 485.07$ &
$\mathrm{NaN} \pm \mathrm{NaN}$
\\
128 & 128 &
$182.36 \pm 33.54$ &
$65.58 \pm 0.26$ &
$2147.13 \pm 538.70$ &
$\mathrm{NaN} \pm \mathrm{NaN}$
\\
      \bottomrule
    \end{tabular}
  \end{small}
\end{table}

\begin{table}[bt]
  \caption{Detailed results of the hyperparameter study for MP-PDE\@. All computations are done on the same CPU (Intel Xeon CPU E5-2695 v2 @ 2.40 GHz) with one core.}
  \centering
  \label{tab:mppde_hyper}
  \begin{small}
    \begin{tabular}{rrrrrrr}
      \toprule
      $n_\mathrm{bundle}$
      &
      $n_\mathrm{feature}$
      &
      $n_\mathrm{neighbor}$
      &
      \makecell{%
        Computation time
        \\
        on the test dataset [s]
      }
      &
      \makecell{%
        Loss
        \\
        on the test dataset
        \\
        ($\times 10^{-1}$)
      }
      &
      \makecell{%
        Computation time
        \\
        on the taller dataset [s]
      }
      &
      \makecell{%
        Loss
        \\
        on the taller dataset
        \\
        ($\times 10^{-1}$)
      }
      \\
      \midrule
2 & 32 & 4 &
$0.88 \pm 0.03$ &
$5.12 \pm 0.03$ &
$1.74 \pm 0.02$ &
$18.29 \pm 0.14$
\\
2 & 32 & 8 &
$4.06 \pm 0.13$ &
$4.56 \pm 0.02$ &
$9.81 \pm 0.15$ &
$20.56 \pm 0.09$
\\
2 & 32 & 16 &
$10.80 \pm 0.27$ &
$4.89 \pm 0.02$ &
$27.00 \pm 0.23$ &
$21.06 \pm 0.11$
\\
2 & 64 & 4 &
$1.91 \pm 0.06$ &
$4.26 \pm 0.02$ &
$3.68 \pm 0.03$ &
$20.74 \pm 0.09$
\\
2 & 64 & 8 &
$8.04 \pm 0.22$ &
$4.16 \pm 0.02$ &
$24.31 \pm 0.25$ &
$18.51 \pm 0.09$
\\
2 & 64 & 16 &
$27.75 \pm 0.31$ &
$4.75 \pm 0.02$ &
$56.54 \pm 0.42$ &
$8.97 \pm 0.05$
\\
2 & 128 & 4 &
$4.99 \pm 0.06$ &
$3.88 \pm 0.02$ &
$10.64 \pm 0.15$ &
$14.23 \pm 0.07$
\\
2 & 128 & 8 &
$26.11 \pm 0.32$ &
$3.76 \pm 0.02$ &
$54.61 \pm 0.60$ &
$17.62 \pm 0.10$
\\
2 & 128 & 16 &
$62.50 \pm 0.69$ &
$4.16 \pm 0.02$ &
$127.80 \pm 1.03$ &
$14.56 \pm 0.08$
\\
4 & 32 & 4 &
$0.47 \pm 0.01$ &
$5.31 \pm 0.03$ &
$0.98 \pm 0.01$ &
$10.58 \pm 0.06$
\\
4 & 32 & 8 &
$2.37 \pm 0.05$ &
$4.52 \pm 0.02$ &
$5.28 \pm 0.02$ &
$20.70 \pm 0.10$
\\
4 & 32 & 16 &
$5.87 \pm 0.11$ &
$4.86 \pm 0.02$ &
$14.21 \pm 0.15$ &
$14.68 \pm 0.08$
\\
4 & 64 & 4 &
$0.96 \pm 0.01$ &
$4.08 \pm 0.02$ &
$1.96 \pm 0.03$ &
$10.32 \pm 0.05$
\\
4 & 64 & 8 &
$5.13 \pm 0.07$ &
$3.90 \pm 0.02$ &
$12.73 \pm 0.11$ &
$10.10 \pm 0.05$
\\
4 & 64 & 16 &
$14.39 \pm 0.16$ &
$4.56 \pm 0.02$ &
$28.64 \pm 0.29$ &
$8.95 \pm 0.05$
\\
4 & 128 & 4 &
$2.56 \pm 0.03$ &
$3.49 \pm 0.02$ &
$5.61 \pm 0.02$ &
$13.67 \pm 0.09$
\\
4 & 128 & 8 &
$13.50 \pm 0.17$ &
$3.61 \pm 0.02$ &
$27.88 \pm 0.27$ &
$9.26 \pm 0.06$
\\
4 & 128 & 16 &
$31.97 \pm 0.35$ &
$4.31 \pm 0.02$ &
$65.27 \pm 0.51$ &
$9.73 \pm 0.05$
\\
8 & 32 & 4 &
$0.29 \pm 0.00$ &
$5.66 \pm 0.03$ &
$0.61 \pm 0.01$ &
$11.89 \pm 0.06$
\\
8 & 32 & 8 &
$1.43 \pm 0.02$ &
$4.63 \pm 0.02$ &
$2.37 \pm 0.07$ &
$7.78 \pm 0.04$
\\
8 & 32 & 16 &
$3.36 \pm 0.07$ &
$4.86 \pm 0.02$ &
$8.23 \pm 0.06$ &
$12.84 \pm 0.06$
\\
8 & 64 & 4 &
$0.56 \pm 0.01$ &
$4.57 \pm 0.02$ &
$1.13 \pm 0.02$ &
$8.35 \pm 0.05$
\\
8 & 64 & 8 &
$2.75 \pm 0.04$ &
$4.08 \pm 0.02$ &
$6.76 \pm 0.06$ &
$7.68 \pm 0.05$
\\
8 & 64 & 16 &
$7.35 \pm 0.10$ &
$4.56 \pm 0.02$ &
$15.65 \pm 0.13$ &
$10.88 \pm 0.05$
\\
8 & 128 & 4 &
$1.37 \pm 0.02$ &
$3.77 \pm 0.02$ &
$3.02 \pm 0.03$ &
$7.76 \pm 0.04$
\\
8 & 128 & 8 &
$6.98 \pm 0.09$ &
$3.64 \pm 0.02$ &
$14.33 \pm 0.16$ &
$8.11 \pm 0.04$
\\
8 & 128 & 16 &
$16.31 \pm 0.19$ &
$4.04 \pm 0.02$ &
$33.84 \pm 0.28$ &
$7.46 \pm 0.04$
\\
     \bottomrule
    \end{tabular}
  \end{small}
\end{table}

\begin{table}[bt]
  \caption{Detailed results of the hyperparameter study for PENN\@. All computations are done on the same CPU (Intel Xeon CPU E5-2695 v2 @ 2.40 GHz) with one core.}
  \centering
  \label{tab:penn_hyper}
  \begin{small}
    \begin{tabular}{rrrrrr}
      \toprule
      $n_\mathrm{feature}$
      &
      $n_\mathrm{rep}$
      &
      \makecell{%
        Computation time
        \\
        on the test dataset [s]
      }
      &
      \makecell{%
        Loss
        \\
        on the test dataset
        \\
        ($\times 10^{-1}$)
      }
      &
      \makecell{%
        Computation time
        \\
        on the taller dataset [s]
      }
      &
      \makecell{%
        Loss
        \\
        on the taller dataset
        \\
        ($\times 10^{-1}$)
      }
      \\
      \midrule
4 & 4 &
$5.87 \pm 0.04$ &
$3.80 \pm 0.02$ &
$10.54 \pm 0.07$ &
$8.47 \pm 0.07$
\\
4 & 8 &
$10.43 \pm 0.08$ &
$4.36 \pm 0.02$ &
$18.75 \pm 0.13$ &
$123.80 \pm 1.43$
\\
4 & 16 &
$11.37 \pm 0.55$ &
$6.57 \pm 0.03$ &
$31.15 \pm 0.79$ &
$172376.18 \pm 35812.77$
\\
8 & 4 &
$7.67 \pm 0.06$ &
$2.49 \pm 0.01$ &
$14.47 \pm 0.11$ &
$3.48 \pm 0.02$
\\
8 & 8 &
$13.72 \pm 0.12$ &
$4.13 \pm 0.02$ &
$26.18 \pm 0.19$ &
$4.52 \pm 0.02$
\\
8 & 16 &
$18.91 \pm 0.62$ &
$3.83 \pm 0.02$ &
$47.34 \pm 0.98$ &
$26.96 \pm 0.34$
\\
16 & 4 &
$11.76 \pm 0.10$ &
$2.12 \pm 0.01$ &
$22.95 \pm 0.18$ &
$29.93 \pm 0.87$
\\
16 & 8 &
$21.07 \pm 0.20$ &
$3.57 \pm 0.02$ &
$41.53 \pm 0.32$ &
$28.90 \pm 0.63$
\\
16 & 16 &
$39.59 \pm 0.39$ &
$190.58 \pm 2.09$ &
$78.26 \pm 0.69$ &
$225.39 \pm 1.59$
\\
32 & 4 &
$19.36 \pm 0.19$ &
$1.58 \pm 0.01$ &
$39.21 \pm 0.32$ &
$3.72 \pm 0.02$
\\
32 & 8 &
$34.52 \pm 0.34$ &
$127.20 \pm 0.80$ &
$69.15 \pm 0.62$ &
$200.11 \pm 3.77$
\\
32 & 16 &
$65.37 \pm 0.62$ &
$97.67 \pm 0.42$ &
$130.88 \pm 1.13$ &
$1333.77 \pm 16.01$
\\
64 & 4 &
$34.36 \pm 0.32$ &
$6.78 \pm 0.03$ &
$70.49 \pm 0.60$ &
$85.39 \pm 1.28$
\\
64 & 8 &
$60.68 \pm 0.63$ &
$775.32 \pm 4.13$ &
$125.40 \pm 0.80$ &
$734.80 \pm 5.41$
\\
64 & 16 &
$114.56 \pm 1.10$ &
$150.98 \pm 0.75$ &
$235.70 \pm 1.71$ &
$668.38 \pm 12.94$
\\
      \bottomrule
    \end{tabular}
  \end{small}
\end{table}

\begin{table}[bt]
  \caption{Detailed results of the hyperparameter study for FluxGNN\@. All computations are done on the same CPU (Intel Xeon CPU E5-2695 v2 @ 2.40 GHz) with one core.}
  \centering
  \label{tab:fluxgnn_hyper}
  \begin{small}
    \begin{tabular}{rrrrrrr}
      \toprule
      $n_\mathrm{bundle}$
      &
      $n_\mathrm{feature}$
      &
      $n_\mathrm{rep}$
      &
      \makecell{%
        Computation time
        \\
        on the test dataset [s]
      }
      &
      \makecell{%
        Loss
        \\
        on the test dataset
        \\
        ($\times 10^{-1}$)
      }
      &
      \makecell{%
        Computation time
        \\
        on the taller dataset [s]
      }
      &
      \makecell{%
        Loss
        \\
        on the taller dataset
        \\
        ($\times 10^{-1}$)
      }
      \\
      \midrule
2 & 8 & 2 &
$2.28 \pm 0.04$ &
$3.84 \pm 0.02$ &
$4.38 \pm 0.07$ &
$3.42 \pm 0.02$
\\
2 & 8 & 4 &
$3.06 \pm 0.05$ &
$3.28 \pm 0.02$ &
$5.66 \pm 0.13$ &
$2.91 \pm 0.02$
\\
2 & 8 & 8 &
$4.47 \pm 0.05$ &
$6.81 \pm 0.04$ &
$8.31 \pm 0.24$ &
$5.66 \pm 0.04$
\\
2 & 16 & 2 &
$2.52 \pm 0.04$ &
$4.40 \pm 0.03$ &
$4.85 \pm 0.06$ &
$3.85 \pm 0.03$
\\
2 & 16 & 4 &
$3.55 \pm 0.06$ &
$2.90 \pm 0.01$ &
$6.65 \pm 0.14$ &
$2.67 \pm 0.02$
\\
2 & 16 & 8 &
$5.62 \pm 0.07$ &
$7.91 \pm 0.04$ &
$10.47 \pm 0.14$ &
$6.32 \pm 0.04$
\\
2 & 32 & 2 &
$3.09 \pm 0.05$ &
$3.59 \pm 0.02$ &
$5.87 \pm 0.10$ &
$3.36 \pm 0.02$
\\
2 & 32 & 4 &
$4.60 \pm 0.07$ &
$2.78 \pm 0.01$ &
$8.77 \pm 0.13$ &
$2.83 \pm 0.02$
\\
2 & 32 & 8 &
$7.40 \pm 0.06$ &
$5.58 \pm 0.03$ &
$14.66 \pm 0.18$ &
$5.51 \pm 0.05$
\\
2 & 64 & 2 &
$4.14 \pm 0.07$ &
$3.81 \pm 0.03$ &
$8.11 \pm 0.13$ &
$3.68 \pm 0.03$
\\
2 & 64 & 4 &
$6.67 \pm 0.11$ &
$2.55 \pm 0.01$ &
$13.15 \pm 0.21$ &
$2.66 \pm 0.02$
\\
2 & 64 & 8 &
$10.42 \pm 0.17$ &
$3.22 \pm 0.02$ &
$22.16 \pm 0.18$ &
$3.71 \pm 0.02$
\\
4 & 8 & 2 &
$2.98 \pm 0.06$ &
$4.84 \pm 0.02$ &
$5.82 \pm 0.22$ &
$3.52 \pm 0.02$
\\
4 & 8 & 4 &
$4.21 \pm 0.05$ &
$2.69 \pm 0.01$ &
$7.82 \pm 0.13$ &
$2.49 \pm 0.01$
\\
4 & 8 & 8 &
$7.44 \pm 0.06$ &
$9.20 \pm 0.04$ &
$13.93 \pm 0.30$ &
$6.66 \pm 0.04$
\\
4 & 16 & 2 &
$3.46 \pm 0.07$ &
$7.63 \pm 0.04$ &
$6.75 \pm 0.24$ &
$5.29 \pm 0.05$
\\
4 & 16 & 4 &
$5.72 \pm 0.12$ &
$2.86 \pm 0.01$ &
$11.18 \pm 0.62$ &
$2.93 \pm 0.02$
\\
4 & 16 & 8 &
$8.69 \pm 0.16$ &
$2.74 \pm 0.01$ &
$17.05 \pm 0.34$ &
$2.58 \pm 0.01$
\\
4 & 32 & 2 &
$4.24 \pm 0.05$ &
$3.03 \pm 0.01$ &
$8.31 \pm 0.09$ &
$2.71 \pm 0.02$
\\
4 & 32 & 4 &
$6.80 \pm 0.09$ &
$2.72 \pm 0.02$ &
$13.49 \pm 0.05$ &
$3.18 \pm 0.03$
\\
4 & 32 & 8 &
$12.43 \pm 0.14$ &
$2.60 \pm 0.01$ &
$26.06 \pm 0.19$ &
$2.87 \pm 0.02$
\\
4 & 64 & 2 &
$6.37 \pm 0.08$ &
$2.97 \pm 0.01$ &
$12.49 \pm 0.19$ &
$2.73 \pm 0.02$
\\
4 & 64 & 4 &
$10.84 \pm 0.15$ &
$2.78 \pm 0.02$ &
$21.44 \pm 0.14$ &
$3.13 \pm 0.02$
\\
4 & 64 & 8 &
$21.03 \pm 0.23$ &
$3.64 \pm 0.02$ &
$46.46 \pm 0.88$ &
$3.75 \pm 0.02$
\\
      \bottomrule
    \end{tabular}
  \end{small}
\end{table}

\clearpage
\subsection{Longer Temporal Rollout}
\label{app:rollout}
\begin{figure}[tb]
  \centering
  \centerline{\includegraphics[trim={0cm 0cm 10cm 0cm},width=0.95\columnwidth]{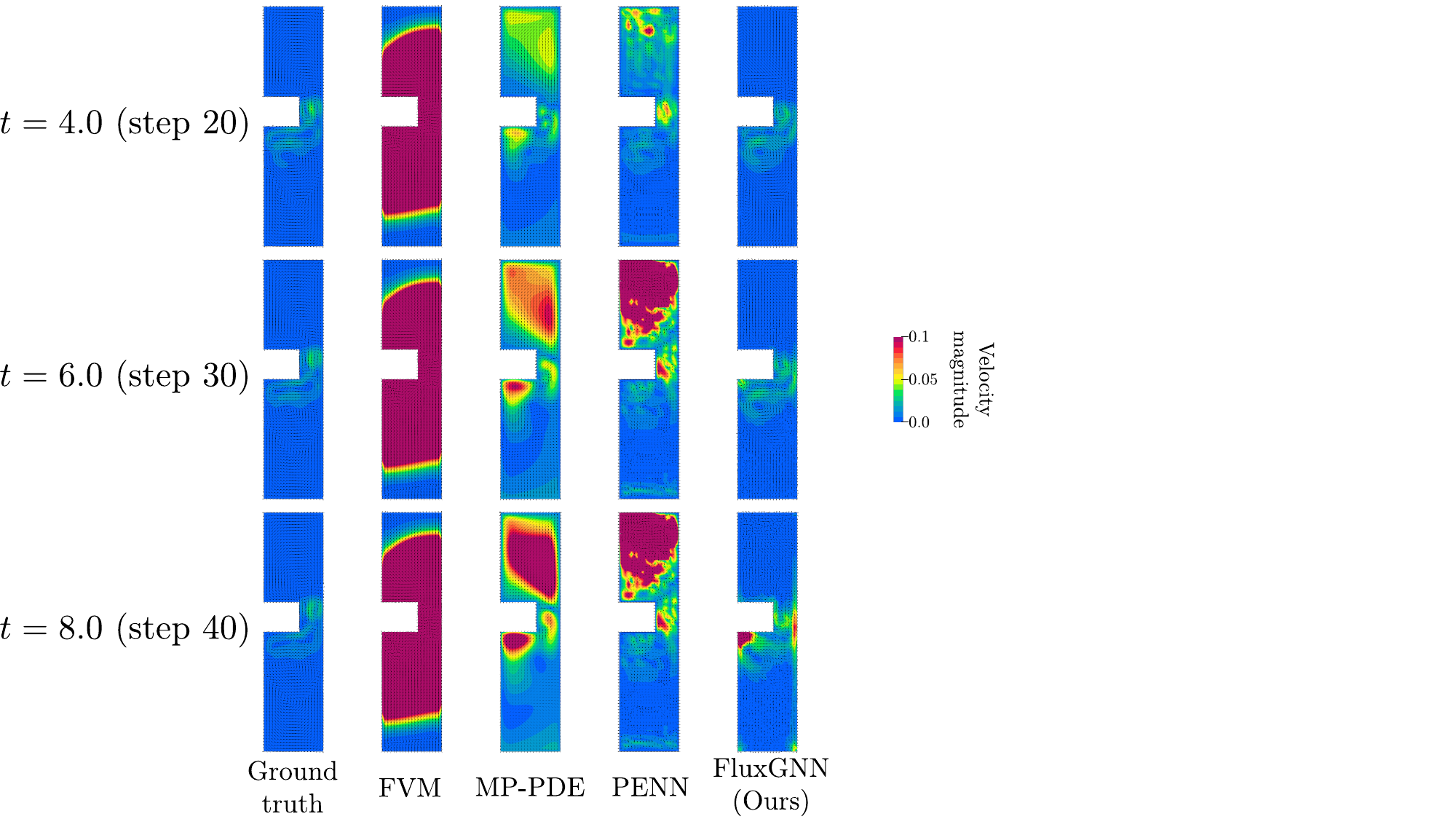}}
    \caption{Time evolution of velocity field for ground truth, FVM, MP-PDE, PENN, and FluxGNN.}
  \label{fig:ts_u}
\end{figure}

\begin{figure}[tb]
  \centering
  \centerline{\includegraphics[trim={0cm 0cm 10cm 0cm},width=0.95\columnwidth]{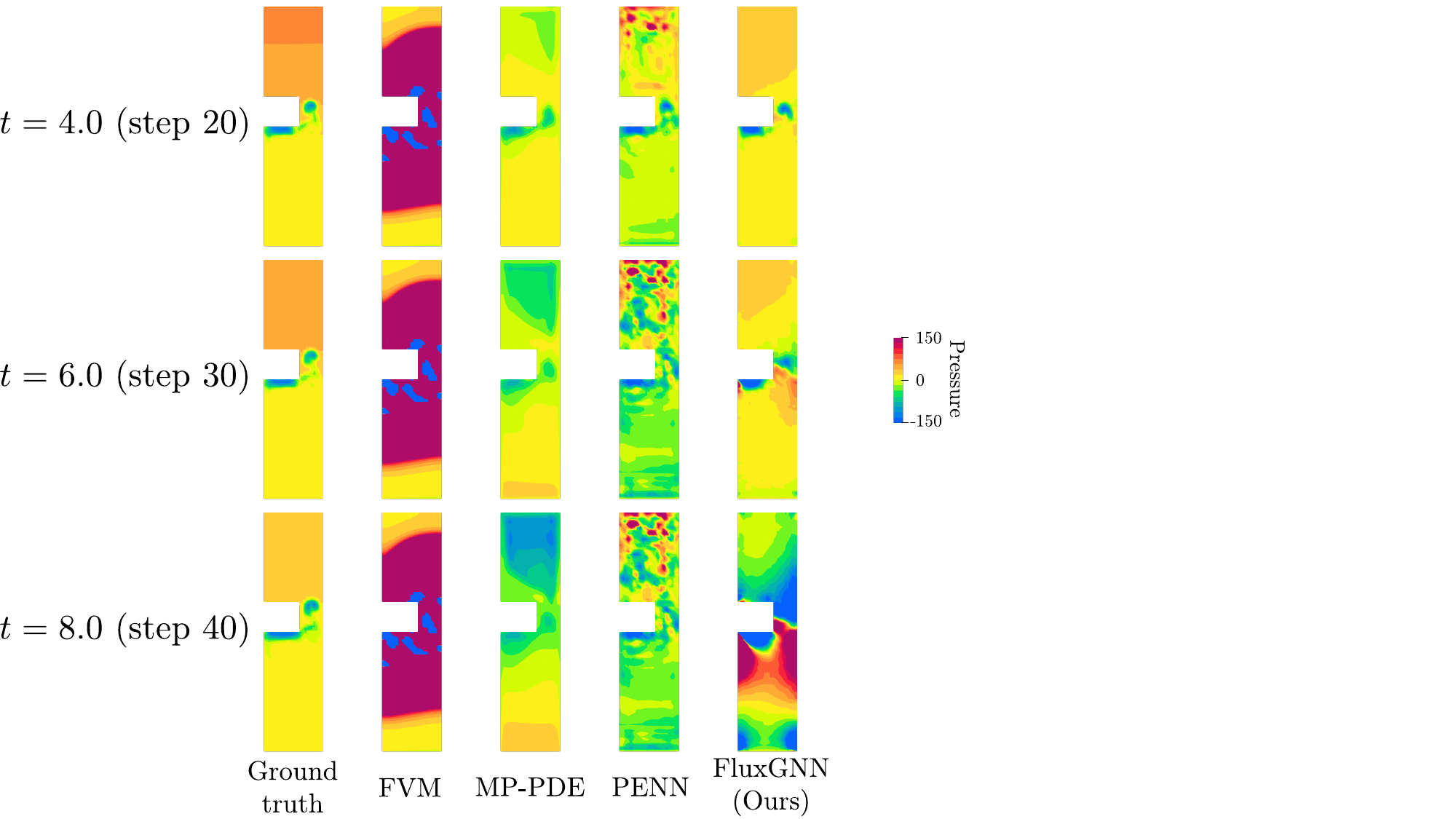}}
    \caption{Time evolution of pressure field for ground truth, FVM, MP-PDE, PENN, and FluxGNN.}
  \label{fig:ts_p}
\end{figure}

\begin{figure}[tb]
  \centering
  \centerline{\includegraphics[trim={0cm 0cm 10cm 0cm},width=0.95\columnwidth]{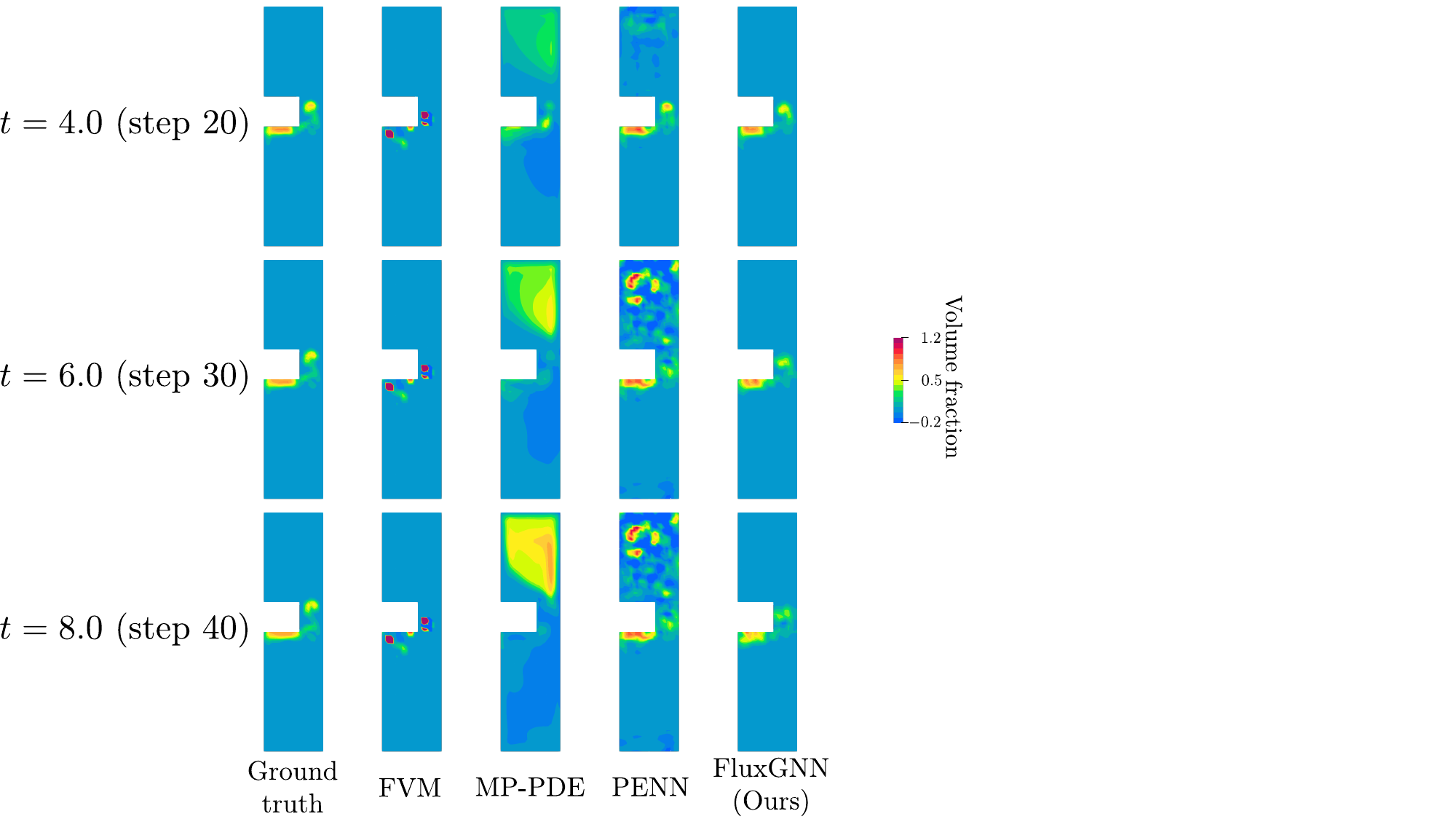}}
    \caption{Time evolution of volume fraction field for ground truth, FVM, MP-PDE, PENN, and FluxGNN.}
  \label{fig:ts_alpha}
\end{figure}

\cref{fig:ts_u,fig:ts_p,fig:ts_alpha} visualize velocity, pressure, and volume fraction fields of longer temporal rollout predictions. Our observations are as follows:
\begin{itemize}
  \item
    A substantial part of the error of FluxGNN is mainly due to the pressure. It seems that pressure starts to be unstable earlier than velocity and volume fraction, implying that instability of pressure would be key to establishing a more stable model. MP-PDE has a lower loss than FluxGNN at $t$ ~ 8.0, but these outputs are not qualitatively similar to those of ground truth.
  \item
    FluxGNN keeps the distribution of the volume fraction field ($\alpha$) similar to ground truth, while other methods show entirely dissimilar fields. That may be due to the conservative property of our method because the volume fraction is strictly conservative in the present setting.
\end{itemize}

\clearpage

\clearpage

\clearpage
\subsection{Ablation Study}
\label{app:ablation}

\cref{tab:ablation} presents the results of an ablation study. That suggests that all components included in the model contribute to spatial out-of-domain generalizability. We compared the following models:
\begin{itemize}
  \item
    Without temporal bundling
  \item
    Without neural nonlinear solver, meaning explicit temporal differentiation
  \item
    Without conservation, meaning no permutation-invariance in the flux functions
  \item
    Without scaling-equivariance
  \item
    Without $\mathrm{E}(n)$-equivariance
  \item
    FluxGNN, the proposed method
\end{itemize}

The model without $\mathrm{E}(n)$-equivariance performs best on the test, rotation, and scaling datasets. In particular, it performs well on the rotation dataset, although the model has no $\mathrm{E}(n)$-equivariance. That implies that the inductive bias, introduced with an FVM-like computation procedure, works quite fine.

FluxGNN performs the best on the taller dataset. The model without $\mathrm{E}(n)$-equivariance degrades on the taller dataset, implying insufficient generalizability. Our primal goal is to construct reliable methods toward out-of-domain generalizability. Therefore, all the suggested components are necessary to achieve our goal.

\begin{table}[tbh]
  \caption{RMSE loss and conservation error ($\pm$ standard error of the mean) on the evaluation datasets of the Navier--Stokes equation with mixture for an ablation study. Each metric is normalized using standard deviation.}
  \centering
  \label{tab:ablation}
  \begin{small}
    \begin{tabular}{lrrrrr}
      \toprule
      Method
      &
      Dataset
      &
      Loss $\vu$ $\left(\times 10^{-1} \right)$
      &
      Loss $p$ $\left(\times 10^{-1} \right)$
      &
      Loss $\alpha$ $\left(\times 10^{-1} \right)$
      &
      Conservation error $\alpha$ $\left(\times 10^{-5} \right)$
      \\
      \midrule
        w/o temporal bundling & test & 2.075 $\pm$ 0.015 & 5.209 $\pm$ 0.034 & 2.851 $\pm$ 0.030 & \textbf{0.01} $\pm$ 0.00 \\ 
        w/o neural nonlinear solver & test & 1.291 $\pm$ 0.009 & 1.791 $\pm$ 0.011 & 0.429 $\pm$ 0.005 & 0.91 $\pm$ 0.33 \\ 
        w/o conservation & test & 1.252 $\pm$ 0.009 & 1.381 $\pm$ 0.014 & 0.405 $\pm$ 0.006 & 86.89 $\pm$ 3.88 \\ 
        w/o scaling-equivariance & test & 1.835 $\pm$ 0.013 & 1.410 $\pm$ 0.010 & 0.492 $\pm$ 0.007 & 0.05 $\pm$ 0.01 \\ 
        w/o $\En$-equivariance & test & \textbf{0.839} $\pm$ 0.005 & 1.192 $\pm$ 0.007 & \textbf{0.282} $\pm$ 0.004 & \textbf{0.01} $\pm$ 0.00 \\ 
        FluxGNN & test & 1.202 $\pm$ 0.008 & \textbf{1.143} $\pm$ 0.008 & 0.349 $\pm$ 0.005 & 0.06 $\pm$ 0.03
        \\[5pt]
        w/o temporal bundling & rotation & 2.114 $\pm$ 0.012 & 3.703 $\pm$ 0.028 & 2.850 $\pm$ 0.030 & \textbf{0.01} $\pm$ 0.00 \\ 
        w/o neural nonlinear solver & rotation & 1.290 $\pm$ 0.007 & 1.859 $\pm$ 0.010 & 0.426 $\pm$ 0.005 & \textbf{0.01} $\pm$ 0.00 \\ 
        w/o conservation & rotation & 1.248 $\pm$ 0.007 & 1.966 $\pm$ 0.014 & 0.414 $\pm$ 0.006 & 86.89 $\pm$ 3.88 \\ 
        w/o scaling-equivariance & rotation & 1.834 $\pm$ 0.010 & 1.620 $\pm$ 0.010 & 0.489 $\pm$ 0.007 & \textbf{0.01} $\pm$ 0.00 \\ 
        w/o $\En$-equivariance & rotation & \textbf{0.852} $\pm$ 0.005 & 1.306 $\pm$ 0.007 & \textbf{0.296} $\pm$ 0.004 & \textbf{0.01} $\pm$ 0.00 \\ 
        FluxGNN & rotation & 1.207 $\pm$ 0.007 & \textbf{1.175} $\pm$ 0.008 & 0.351 $\pm$ 0.005 & \textbf{0.01} $\pm$ 0.00
        \\[5pt]
        w/o temporal bundling & scaling & 2.070 $\pm$ 0.015 & 5.812 $\pm$ 0.037 & 2.850 $\pm$ 0.030 & \textbf{0.01} $\pm$ 0.00 \\ 
        w/o neural nonlinear solver & scaling & 1.316 $\pm$ 0.009 & 1.790 $\pm$ 0.011 & 0.438 $\pm$ 0.005 & 1.80 $\pm$ 0.87 \\ 
        w/o conservation & scaling & 1.245 $\pm$ 0.009 & 1.415 $\pm$ 0.013 & 0.410 $\pm$ 0.006 & 86.90 $\pm$ 3.89 \\ 
        w/o scaling-equivariance & scaling & 1.922 $\pm$ 0.014 & 8.102 $\pm$ 0.782 & 27.797 $\pm$ 3.080 & 0.05 $\pm$ 0.02 \\ 
        w/o $\En$-equivariance & scaling & \textbf{0.839} $\pm$ 0.005 & \textbf{1.189} $\pm$ 0.007 & \textbf{0.280} $\pm$ 0.003 & \textbf{0.01} $\pm$ 0.00 \\ 
        FluxGNN & scaling & 1.219 $\pm$ 0.009 & 1.228 $\pm$ 0.008 & 0.356 $\pm$ 0.005 & 0.05 $\pm$ 0.01
        \\[5pt]
        w/o temporal bundling & taller & 1.822 $\pm$ 0.017 & 4.329 $\pm$ 0.029 & 2.212 $\pm$ 0.034 & \textbf{0.01} $\pm$ 0.00 \\
        w/o neural nonlinear solver & taller & 1.268 $\pm$ 0.010 & 1.202 $\pm$ 0.010 & 0.403 $\pm$ 0.007 & 0.34 $\pm$ 0.05 \\
        w/o conservation & taller & 1.269 $\pm$ 0.012 & 1.336 $\pm$ 0.012 & 0.473 $\pm$ 0.009 & 205.37 $\pm$ 90.01 \\
        w/o scaling-equivariance & taller & 1.604 $\pm$ 0.015 & 1.102 $\pm$ 0.009 & 0.398 $\pm$ 0.007 & 0.03 $\pm$ 0.00 \\
        w/o $\En$-equivariance & taller & \textbf{1.122} $\pm$ 0.009 & 1.129 $\pm$ 0.009 & 0.440 $\pm$ 0.008 & \textbf{0.01} $\pm$ 0.00 \\
        FluxGNN & taller & 1.184 $\pm$ 0.009 & \textbf{0.966} $\pm$ 0.008 & \textbf{0.337} $\pm$ 0.006 & 0.02 $\pm$ 0.00 \\
      \bottomrule
    \end{tabular}
  \end{small}
\end{table}

\end{document}